\documentclass[sigconf]{acmart}
\newcommand{\proposed}{\textsf{RGRL}}
\usepackage[linesnumbered,ruled]{algorithm2e}
\usepackage{subcaption}
\usepackage{enumitem}
\usepackage{multirow}

\usepackage{color, colortbl}
\definecolor{Gray}{gray}{0.9}

\usepackage{xcolor}
\usepackage{pifont}
\usepackage{setspace}
\newcommand{\cmark}{\textcolor{blue}{\ding{51}}}%
\newcommand{\xmark}{\textcolor{red}{\ding{55}}}%

\usepackage{tablefootnote}

\usepackage{balance}

\AtBeginDocument{%
  \providecommand\BibTeX{{%
    \normalfont B\kern-0.5em{\scshape i\kern-0.25em b}\kern-0.8em\TeX}}}

\copyrightyear{2022}
\acmYear{2022}
\setcopyright{acmcopyright}
\acmConference[CIKM '22] {Proceedings of the 31st ACM International Conference on Information and Knowledge Management}{October 17--21, 2022}{Atlanta, GA, USA.}
\acmBooktitle{Proceedings of the 31st ACM International Conference on Information and Knowledge Management (CIKM '22), October 17--21, 2022, Atlanta, GA, USA}
\acmPrice{15.00}
\acmISBN{978-1-4503-9236-5/22/10}
\acmDOI{10.1145/3511808.3557428}

\begin{document}

\title{Relational Self-Supervised Learning on Graphs}




\author{Namkyeong Lee}
\affiliation{%
  \institution{KAIST ISysE}
  \city{Daejeon}
  \country{Republic of Korea}
}
\email{namkyeong96@kaist.ac.kr}

\author{Dongmin Hyun}
\affiliation{%
  \institution{POSTECH PIAI}
  \city{Pohang}
  \country{Republic of Korea}
}
\email{dm.hyun@postech.ac.kr}

\author{Junseok Lee}
\affiliation{%
  \institution{KAIST ISysE}
  \city{Daejeon}
  \country{Republic of Korea}
}
\email{junseoklee@kaist.ac.kr}

\author{Chanyoung Park}
\affiliation{%
  \institution{KAIST ISysE \& AI}
  \city{Daejeon}
  \country{Republic of Korea}
}
\email{cy.park@kaist.ac.kr}
\authornote{
Corresponding author.
}

\renewcommand{\shortauthors}{Namkyeong Lee, Dongmin Hyun, Junseok Lee, \& Chanyoung Park}

\begin{abstract}
    Over the past few years, graph representation learning (GRL) has been a powerful strategy for analyzing graph-structured data. 
    Recently, GRL methods have shown promising results by adopting self-supervised learning methods developed for learning representations of images. Despite their success, existing GRL methods tend to overlook an inherent distinction between images and graphs, i.e., images are assumed to be independently and identically distributed, whereas graphs exhibit relational information among data instances, i.e., nodes.
    To fully benefit from the relational information inherent in the graph-structured data,
    we propose a novel GRL method, called \proposed
    , that learns from the relational information generated from the graph itself.
    ~\proposed~learns node representations such that the relationship among nodes is invariant to augmentations, i.e., \textit{augmentation-invariant relationship}, which allows the node representations to vary as long as the relationship among the nodes is preserved.
    By considering the relationship among nodes in both global and local perspectives,~\proposed~overcomes limitations of previous contrastive and non-contrastive methods, and achieves the best of both worlds. Extensive experiments on fourteen benchmark datasets over various downstream tasks demonstrate the superiority of~\proposed~over state-of-the-art baselines.
    The source code for~\proposed~is available at~\url{https://github.com/Namkyeong/RGRL}.
\end{abstract}

\ccsdesc[500]{Computing methodologies~Machine learning algorithms}
\ccsdesc[500]{Computing methodologies~Unsupervised learning}

\keywords{Self-Supervised Learning, Graph Representation Learning, Graph Neural Networks}


\maketitle

\newpage

\section{Introduction}

\begin{figure}[t]
    \centering
    \includegraphics[width=0.85\columnwidth]{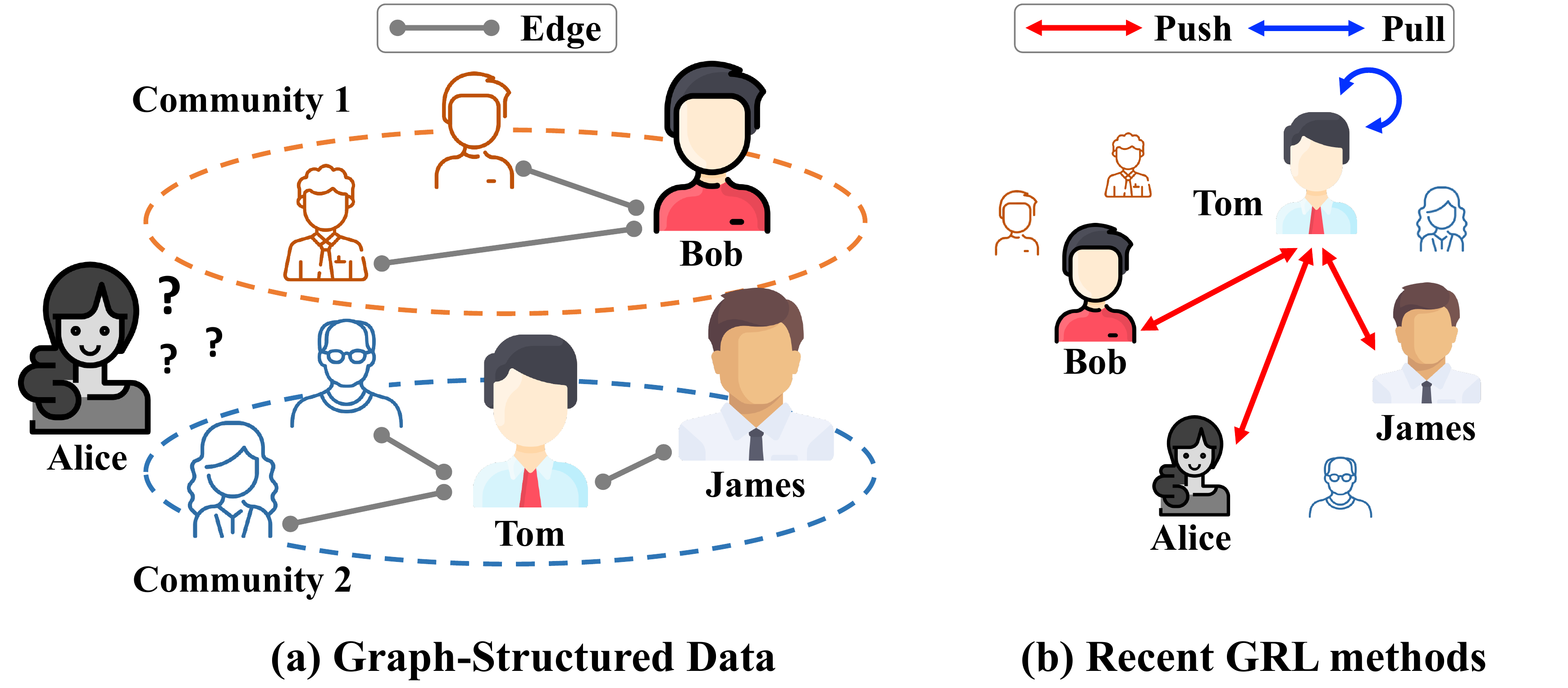} 
    \vspace{-2ex}
    \caption{Recent GRL methods cannot fully benefit from the relational information of given graph-structured data.}
    \label{fig1}
    \vspace{-3ex}
\end{figure}


Recently, self-supervised learning paradigm, which trains models on pretext tasks derived solely from data without any label information, achieved great success in many domains \cite{BERT,GPT3,SimCLR,BYOL}.
Among various self-supervised learning approaches, contrastive learning has shown its effectiveness in representation learning by pulling semantically similar (i.e., positive) pairs of data instances together and pushing dissimilar (i.e., negative) ones apart~\cite{DeepInfomax,SimCLR}. 
In general, successful contrastive methods follow the principle of instance discrimination \cite{IND}, which pairs data instances based on whether they are derived from the same instance (i.e., positive pairs) or not (i.e., negative pairs). 

Inspired by the success of the contrastive methods in computer vision, these methods have been recently adopted to representation learning on graphs. 
However, existing contrastive learning-based GRL methods~\cite{GCA,GRACE,DGI} closely follow the model architectures that were successful on images without considering an inherent distinction between images and graphs, i.e., images are assumed to be independently and identically distributed, whereas \textit{graphs exhibit relational information among nodes}.
For example, GRACE~\cite{GRACE} and GCA~\cite{GCA} inherit the instance discrimination principle~\cite{IND,SimCLR}, and treat all other nodes apart from the node itself as negatives.
However, we argue that without considering the relational information inherent in graphs, these methods are prone to \textit{sampling bias}~\cite{chuang2020debiased,lin2021prototypical,lee2022augmentation}, i.e., some negative samples are in fact semantically similar to the query node.

To illustrate the sampling bias of existing methods, consider Tom as the query node in Fig.~\ref{fig1}. Then, James would be regarded as a negative sample {(Fig.~\ref{fig1}b)} even though James belongs to the same community as Tom {(Fig.~\ref{fig1}a)}, which means that they are likely to share some interest.
To make the matter worse, James would be treated equally as negative to Tom as Bob is to Tom {(Fig.~\ref{fig1}b)}, even though Bob belongs to a different community {(Fig.~\ref{fig1}a)}.
These false supervisory signals can seriously interfere with representation learning on graphs, and eventually degrade the performance on downstream tasks, such as node classification, and link prediction. BGRL~\cite{BGRL}, which is a recent non-contrastive method, avoids the sampling bias by relying only on positive samples, i.e., ``pull'' only in Fig.~\ref{fig1}b.
However, since BGRL is trained by predicting an augmented version of a node itself, it still cannot fully benefit from the relational information inherent in the graph-structured data.
For example, assume that we lack information about Alice (i.e., less informative or noisy features) in Fig.~\ref{fig1}a.
In this case, if we only considered Alice herself along with her augmented version to train her own representation as done in BGRL (i.e., ``pull'' only in Fig.~\ref{fig1}b), it would be non-trivial to discover which community she is likely to join.
{However, the problem can be alleviated by using relational information in the graph. That is, considering Alice's relationship with Bob and Tom would make it easier to discover which community Alice is likely to join.}


To this end, we propose \textsf{R}elational \textsf{G}raph \textsf{R}epresentation \textsf{L}earning (\proposed), a simple yet effective self-supervised learning framework for graphs, that benefits from the relational information inherent in the graph-structured data.
The main idea is to \textit{allow node representations to vary as long as the relationship among the nodes is preserved},
rather than 1) strictly distinguishing positive nodes from negative nodes as done by existing contrastive methods~\cite{GCA,GRACE}, or 2) strictly enforcing the node representations to be augmentation-invariant as done by existing non-contrastive methods~\cite{BGRL, zhang2021canonical}.
More precisely, given two GNN-based encoders each of which encodes an augmented view of a graph, the first encoder calculates the similarity of a query node with respect to a set of sampled anchor nodes, while the second encoder tries to mimic the computed query-anchors similarity. 
By doing so, \proposed~
1) relaxes the binary nature of contrastive methods with soft labeling, thereby alleviating the sampling bias, and 2) learns node representations such that the relationship among nodes is invariant to augmentations, i.e., \textit{augmentation-invariant relationship}, rather than augmentation-invariant node representation~\cite{BGRL,GCA}. 
Consequently, we expect~\proposed~to capture the core relationship among nodes that should be preserved no matter how the graph is perturbed.

A key challenge of~\proposed~is how to sample anchor nodes, as GNN-based models are known to be \textit{degree-biased}~\cite{tang2020investigating}, i.e., nodes with higher degree tend to result in higher quality representations, as shown in Fig.~\ref{fig:degree_bias}. This is caused by the fact that high-degree nodes receive more information during neighborhood aggregation compared with low-degree nodes.
Hence, we introduce a technique to sample anchor nodes so that nodes with lower degree are sampled more frequently than nodes with higher degree. 
By doing so, the parameters of~\proposed~are optimized by focusing more on low-degree nodes, thereby alleviating the degree-biased issue. 
Moreover, besides sampling the anchor nodes in the \textit{global} perspective, we additionally sample nodes that are structurally close to the query node to capture the \textit{local} structural information given in a graph.

Our extensive experiments on \textbf{fourteen} real-world datasets on various downstream tasks, i.e., node classification in both homogeneous and heterogeneous graphs, and link prediction, demonstrate the superiority of \proposed. 
Moreover, our qualitative analysis shows that~\proposed~indeed captures the core relationship among nodes.
Further appeals of~\proposed~are that~\proposed~learns high-quality representations of nodes with 1) less informative input features, which demonstrates the robustness of~\proposed, and nodes with 2) low degree, both of which demonstrate the practicality of~\proposed~in real-world applications.

\vspace{-1.5ex}
\section{Related Work}
\subsection{Graph Representation Learning}
\vspace{-0.5ex}
Recent years have seen a surge of interest in analyzing graph-structured data by using various machine learning approaches. Inspired by word2vec~\cite{word2vec}, DeepWalk~\cite{deepwalk} and node2vec~\cite{node2vec} generate node sequences through random walks, and apply skip-gram approach to map nodes appearing within the same context into similar vector representations. While these models have shown success in various tasks such as node classification and link prediction, they overemphasize the proximity information at the expense of structural information \cite{DGI}, and cannot incorporate node attributes~\cite{GCN,DGI}. Graph Neural Networks (GNNs)~\cite{GCN,GAT,Graphsage} address these limitations by learning node representations through recursively aggregating the features of neighboring nodes. 
However, GNNs still require a sufficient number of labeled data for training, which is impractical in reality~\cite{DGI}.

\begin{figure}[t]
    \centering
    \includegraphics[width=0.65\linewidth]{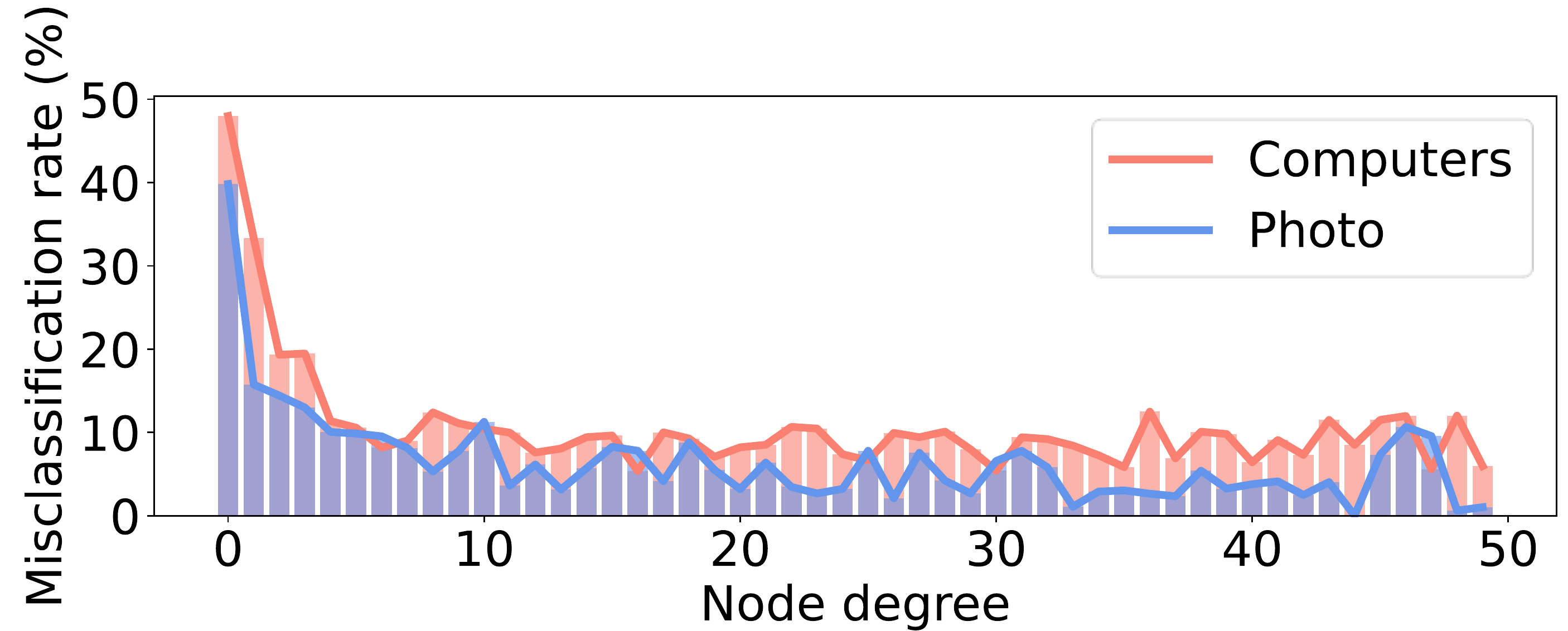} 
    \vspace{-2ex}
    \caption{GNN-based models are biased towards high-degree nodes. Low-degree nodes tend to be misclassified.}
    \label{fig:degree_bias}
\end{figure}

\vspace{-1ex}
\subsection{Self-Supervised Learning on Graphs}
\noindent\textbf{Contrastive learning-based.}
Motivated by the great success of contrastive methods in computer vision applied on images, these methods have recently been adopted to graphs~\cite{MVGRL,zhang2021canonical,lee2022grafn}. Inspired by Deep Infomax~\cite{DeepInfomax}, DGI~\cite{DGI} learns node representations by maximizing the mutual information between the local patch of a graph, i.e., node, and the global summary of the graph, thereby capturing the global information of a graph that is overlooked by vanilla graph convolutional networks (GCNs). GRACE \cite{GRACE}, which adopts SimCLR~\cite{SimCLR} to graph domain, learns node representations with two augmented views of a graph. Specifically, GRACE first creates two augmented views of a graph by randomly dropping edges or masking their features. Then, it pulls representations of the same node in the two augmented graphs while pushing apart representations of every other node. GCA \cite{GCA} enhances GRACE by introducing adaptive augmentation techniques that focus on the graph structure.
Despite the success of contrastive methods on graphs, they suffer from the sampling bias issue incurred by false negatives among the negative samples \cite{chuang2020debiased, lin2021prototypical, xia2021debiased}. 
Although recent DGCL \cite{xia2021debiased} alleviates the issue by utilizing the probability as the weight of negative samples, it still treats all other nodes as negative samples, thus cannot fully leverage the relational information inherent in graphs.
Moreover, these methods require high computational cost and memory usage owing to a large amount of negative samples required for the model training~\cite{BGRL}.

\smallskip
\noindent\textbf{Non-contrastive learning-based.}
Recent non-contrastive methods avoid the aforementioned limitations by not using negative samples \cite{BGRL, zhang2021canonical}.
BGRL \cite{BGRL} learns node representations by encoding two augmented versions of a graph using two separate encoders: one is trained by maximizing the cosine similarity between the representations generated by the two encoders, while the other encoder is updated by an exponential moving average of the first encoder.
While following the conventional augmentation-invariant learning scheme, CCA-SSG \cite{zhang2021canonical} learns node representations by incorporating additional decorrelation terms so that each dimension of the node representation captures distinct semantics of the node.
Despite avoiding the sampling bias, non-contrastive learning methods still overlook the relational information among nodes in a graph by treating each node to be independent from other nodes after message passing.
In this work, we propose a general framework for learning node representations that leverages relational information as supervisory signals to learn augmentation-invariant relationship among nodes.



\section{Problem Statement}
\noindent\textbf{Notations.}
Let $\mathcal{G} = (\mathcal{V}, \mathcal{E})$ denote a graph, where $\mathcal{V} = \left \{v_1, ... , v_N \right \}$ represents the set of nodes, and ~$\mathcal{E} \subseteq \mathcal{V} \times \mathcal{V} $ represents the set of edges. $\mathcal{G}$ is associated with a feature matrix $\mathbf{X} \in \mathbb{R}^{N \times F}$, and an adjacency matrix $\mathbf{A} \in \mathbb{R}^{N \times N}$ where $\mathbf{A}_{ij} = 1$ if and only if $(v_i, v_j) \in \mathcal{E}$ and $\mathbf{A}_{ij} = 0$ otherwise. \\ 
\noindent\textbf{Task: Unsupervised Graph Representation Learning.} Given a graph $\mathcal{G}$ along with $\mathbf{X}$ and $\mathbf{A}$, we aim to learn an encoder $f(\cdot)$ that produces node representations $\mathbf{H} = f(\mathbf{X}, \mathbf{A})\in\mathbb{R}^{N\times D}$, where $D \ll F$. Our goal is to learn node representations that generalize well to various downstream tasks without using any labeled data.

\section{Proposed Framework:~\proposed}
\begin{figure}[t]
    \centering
    \includegraphics[width=0.99\linewidth]{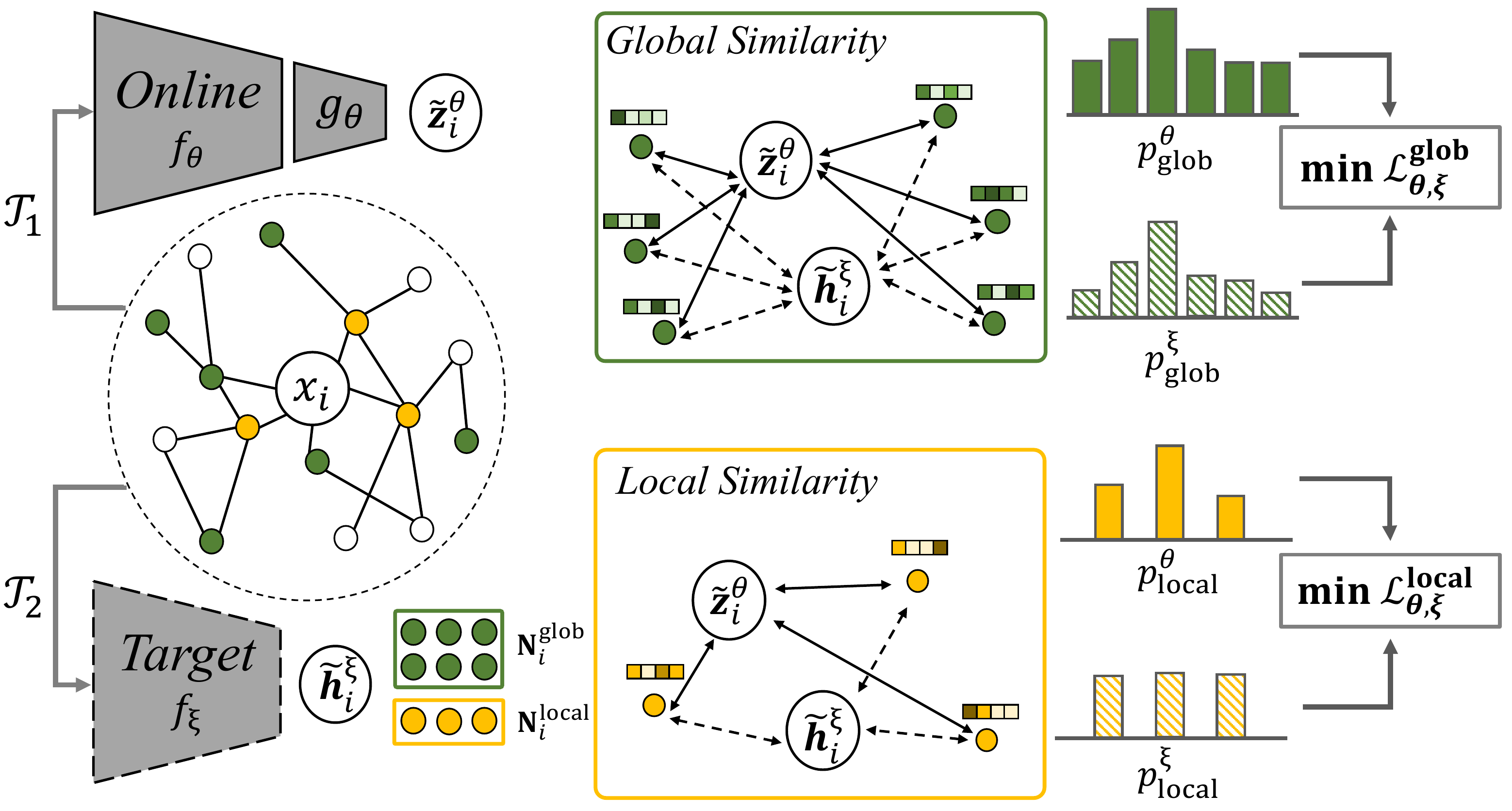} 
    \caption{Overall model architecture of~\proposed.}
    \label{fig:model_architecture}
    \vspace{-2ex}
\end{figure}

In this section,
we first explain how~\proposed~learns the representation of a query node by preserving its similarity with anchor nodes (\textbf{Sec.~\ref{sec:Relational Graph Representation Learning}}) that are sampled regarding both global (\textbf{Sec.~\ref{sec:Capturing global similarity}}) and local (\textbf{Sec.~\ref{sec:Capturing local similarity}}) perspectives.
Then, we explain how the parameters of~\proposed~are updated (\textbf{Sec.~\ref{sec:model update}}).
Lastly, we introduce how~\proposed~can be extended to heterogeneous graphs \textbf{(Sec.~\ref{sec:multiplex})}.
Fig.~\ref{fig:model_architecture} illustrates the overall architecture of \proposed\footnote{\proposed~adopts BYOL~\cite{BYOL} as the backbone of the framework, which is a recently proposed non-contrastive method for image representation learning.}.

\subsection{Relational Graph Representation Learning}
\label{sec:Relational Graph Representation Learning}
In contrast to existing contrastive methods on graphs, \proposed~focuses on preserving the relationship among nodes for learning node representations. 
More precisely, we first generate two graph views $\Tilde{\mathcal{G}}_{1} = (\mathbf{\Tilde{X}}_1, \mathbf{\Tilde{A}}_1)$ and $\Tilde{\mathcal{G}}_{2} = (\mathbf{\Tilde{X}}_2, \mathbf{\Tilde{A}}_2)$ by applying a stochastic graph augmentation function $\mathcal{T}_{1}$ and $\mathcal{T}_{2}$ to the original graph $\mathcal{G}$, respectively. 
Then, the online encoder $f_{\theta}$ produces online representation $\mathbf{\tilde{H}}^{\theta} = f_{\theta}(\mathbf{\Tilde{X}}_1, \mathbf{\Tilde{A}}_1)\in\mathbb{R}^{N\times D}$, while the target encoder $f_{\xi}$ produces target representation $\mathbf{\Tilde{H}}^{\xi} = f_{\xi}(\mathbf{\Tilde{X}}_2, \mathbf{\Tilde{A}}_2)\in\mathbb{R}^{N\times D}$. The online representation $\mathbf{\tilde{H}}^{\theta}$ is additionally fed into a node-level predictor $g_{\theta}$ to obtain the prediction of the target representation, i.e.,
$\mathbf{\Tilde{Z}}^{\theta} = g_{\theta}(\mathbf{\tilde{H}}^{\theta})\in\mathbb{R}^{N\times {D}}$.


Given a query node $v_i\in\mathcal{V}$, we first sample a set of anchor nodes $\mathbf{N}_i$ from graph $\mathcal{G}$. 
Then, the similarity is calculated between the target embedding $\mathbf{\tilde{h}}^{\xi}_i$ of the query node $v_i$, and the target embeddings $\mathbf{\tilde{h}}_{j}^{\xi}$ of anchor nodes $v_j\in \mathbf{N}_i$, divided by a temperature hyperparameter $\tau_\xi$. The similarity is then converted into a probability distribution via softmax :
\begin{eqnarray}
\small
	p_{i}^{\xi}(j) = \frac{\exp(\mathtt{sim}(\mathbf{\tilde{h}}_{i}^{\xi}, \mathbf{\tilde{h}}_{j}^{\xi})/\tau_{\xi})}{\sum_{k \in \mathbf{N}_i}\exp(\mathtt{sim}(\mathbf{\tilde{h}}_{i}^{\xi}, \mathbf{\tilde{h}}_{k}^{\xi})/\tau_{\xi})}, \forall v_j \in \mathbf{N}_i
	\label{eqn:sim1}
\end{eqnarray}
where $p_{i}^{\xi}(j)\in\mathbb{R}$ is the $j$-th element of $p_{i}^{\xi}\in\mathbb{R}^{|\mathbf{N}_i|}$, and $\mathtt{sim}(\cdot, \cdot)$ denotes the cosine similarity between the two input representations.

Likewise, we calculate the similarity between the online prediction $\mathbf{\tilde{z}}^{\theta}_i$ of the query node $v_i$, and the target embeddings $\mathbf{\tilde{h}}_{j}^{\xi}$ of anchor nodes $v_j\in \mathbf{N}_i$ as follows:

\begin{eqnarray}
\small
	p_{i}^{\theta}(j) = \frac{\exp(\mathtt{sim}(\mathbf{\tilde{z}}_i^{\theta}, \mathbf{\tilde{h}}_{j}^{\xi})/\tau_{\theta})}{\sum_{k \in \mathbf{N}_i}\exp(\mathtt{sim}(\mathbf{\tilde{z}}_{i}^{\theta}, \mathbf{\tilde{h}}_{k}^{\xi})/\tau_{\theta})} ,\forall v_j \in \mathbf{N}_i
	\label{eqn:sim2}
\end{eqnarray}
where $p_{i}^{\theta}(j)$ is the $j$-th element of $p_{i}^{\theta}\in\mathbb{R}^{|\mathbf{N}_i|}$, and $\tau_{\theta}$ is the temperature hyperparameter of online network.

Having computed two probability distributions for each node $v_i\in\mathcal{V}$, i.e., $p_i^\theta$ and $p_i^\xi$, derived from the query-anchors similarity, we minimize the following sum of KL divergence: 
\begin{eqnarray}
\small
	\mathcal{L}_{\theta, \xi} = \sum_{v_i\in\mathcal{V}}KL(p_i^{\theta}~||~p_i^{\xi}).
	\label{eqn:kld}
\end{eqnarray}
By minimizing the above loss, the online network is trained to mimic the relational information captured by the target encoder,
thereby learning node representations such that the relationship among nodes is invariant to augmentations, i.e., \textit{augmentation-invariant relationship}, which is the core relationship among nodes that should be preserved no matter how the graph is perturbed.
Note that the role of the temperature hyperparameters, i.e., $\tau_\xi$ and $\tau_\theta$, is to control the discriminability among the sampled anchor nodes in the representation space by determining the sharpness of the distribution~\cite{zheng2021ressl}. 
That is, given that $\tau_\theta$ is fixed, as $\tau_\xi$ becomes smaller, the target distribution becomes more sharpened, providing a discriminative guidance to the online network. 

\subsection{Sampling Anchor Nodes}
A key challenge of~\proposed~is how to sample anchor nodes for each node $v_i\in\mathcal{V}$, i.e., $\mathbf{N}_i$. 
We argue that diverse relational information regarding both global and local perspectives should be considered. Thus, we sample two distinct sets of anchor nodes $\mathbf{N}_i^{\text{glob}}$ (\textbf{Sec.~\ref{sec:Capturing global similarity}}) and $\mathbf{N}_i^{\text{local}}$ (\textbf{Sec.~\ref{sec:Capturing local similarity}}), that play different roles. 
Specifically, the global anchor nodes $\mathbf{N}_i^{\text{glob}}$ are utilized to learn diverse relationship among the nodes considering the overall graph structure, while the local anchor nodes $\mathbf{N}_i^{\text{local}}$ are leveraged to learn the local fine-grained relationship among the nodes that are structurally close in the graph.


\subsubsection{\textbf{Capturing global similarity}}
\label{sec:Capturing global similarity}
To capture the similarity among the nodes in the global perspective, we can na\"ively sample anchor nodes uniformly from the entire graph. However, such a uniform sampling strategy overlooks the highly-skewed node degree distribution, which is not desired since the quality of node representations is closely related to the node degree. To verify this, we conduct an empirical analysis on two real-world graphs, i.e., Amazon Computers and Amazon Photo~\cite{Amazon}, each of which exhibits a power law distribution of node degree. We perform node classification on node representations trained with the state-of-the-art self-supervised method, i.e., BGRL~\cite{BGRL}. As shown in Fig.~\ref{fig:degree_bias}, we observe that the misclassification rate of low-degree nodes is significantly higher than that of high-degree nodes indicating that the training is biased towards high-degree nodes. 
We attribute this to the neighborhood aggregation scheme of GNNs in that low-degree nodes receive less information compared with high-degree nodes, which leads to underfitting of GNNs to low-degree nodes. 
It is worth noting that our observation aligns with findings of~\cite{tang2020investigating} whose experiments are done under the semi-supervised setting, demonstrating that the \textit{degree-bias} issue is critical for the model performance regardless of the existence of the label information. 
To the best of our knowledge,~\proposed~is the first work that addresses the degree-bias issue in self-supervised learning on graphs.




Based on the above observation, we propose to focus on low-degree nodes while training \proposed. The key idea is to sample a set of anchor nodes $v_j$ from \textit{inverse degree-weighted distribution}, which is designed to sample low-degree nodes more frequently as follows:
\begin{eqnarray}
\small
	w_j = \alpha^{\log (\text{deg}_j + 1)} + \beta
\end{eqnarray}
where $\text{deg}_j$ is the degree of node $v_j$, $\alpha$ and $\beta$ are hyperparameters controlling the skewness of the distribution and the minimum sampling weight, respectively. 
It is worth noting that $\alpha$ is set to a value between 0 and 1 to \textit{approximate the misclassification rate distribution shown in Fig.~\ref{fig:degree_bias}}. This implies that nodes with high misclassification rate will be sampled more frequently thereby alleviating the degree-based issue of GNNs.
Finally, we normalize the sampling weight across the nodes, and assign the sampling probability of node $v_j$ as:
\begin{eqnarray}
\small
	p_{sample}(j) = \frac{w_j}{\sum_{v_k\in\mathcal{V}}{w_k}}\, , \forall v_j \in \mathcal{V}
	\label{eqn:sampling_dist}
\end{eqnarray}
In summary, given a query node $v_i$, we sample a set of anchor nodes $\mathbf{N}_i^{\text{glob}}$ from the distribution defined in Eqn.~\ref{eqn:sampling_dist}. Then, we compute the two similarity distributions defined in Eqn.~\ref{eqn:sim1} and Eqn.~\ref{eqn:sim2}, i.e., $p^{\theta}_{\text{glob}}$ and $p^{\xi}_{\text{glob}}$ followed by their KD divergence loss, i.e., $\mathcal{L}_{\theta, \xi}^{\text{glob}}$, defined in Eqn.~\ref{eqn:kld}.
It is worth noting that the similarity distributions computed in this manner capture the query-anchors similarity in the \textit{global perspective}, since the anchor nodes are sampled from the entire graph regardless of the structural information of the graph. 

\subsubsection{\textbf{Capturing local similarity}}
\label{sec:Capturing local similarity}

{By capturing the similarity among nodes in the global perspective, \proposed~learns diverse relationship among nodes without accounting for the explicit structural information, which is crucial for learning local relationship in graph-structured data. Although the GNN encoders naturally capture the local structural information through the neighborhood aggregation scheme, we argue that GNNs fail to capture fine-grained relationship among nodes, i.e., the relationship among the nodes that belong to the same class.
For example, in an academic publication network, even though authors who belong to ``Data Mining'' class and ``Machine Learning'' class may be structurally close to each other due to their similarity in some research topics, these classes are distinguished from each other in a fine-grained perspective. In other words, authors who belong to ``Data Mining'' class are more similar to each other than to authors who belong to ``Machine Learning''. 
However, since GNNs map nodes that are structurally close to similar representations, authors from the two different classes may be mapped to similar representations. 
Hence, to capture fine-grained relationship among structurally close nodes, we propose to sample anchor nodes that 1) are structurally close with the query node in the graph, and at the same time 2) share the same label information with the query node. Such a set of anchor nodes can be discovered by various approaches, such as adjacency, $K$-NN and diffusion:}

\begin{figure}[t]
    \centering
    \includegraphics[width=0.7\linewidth]{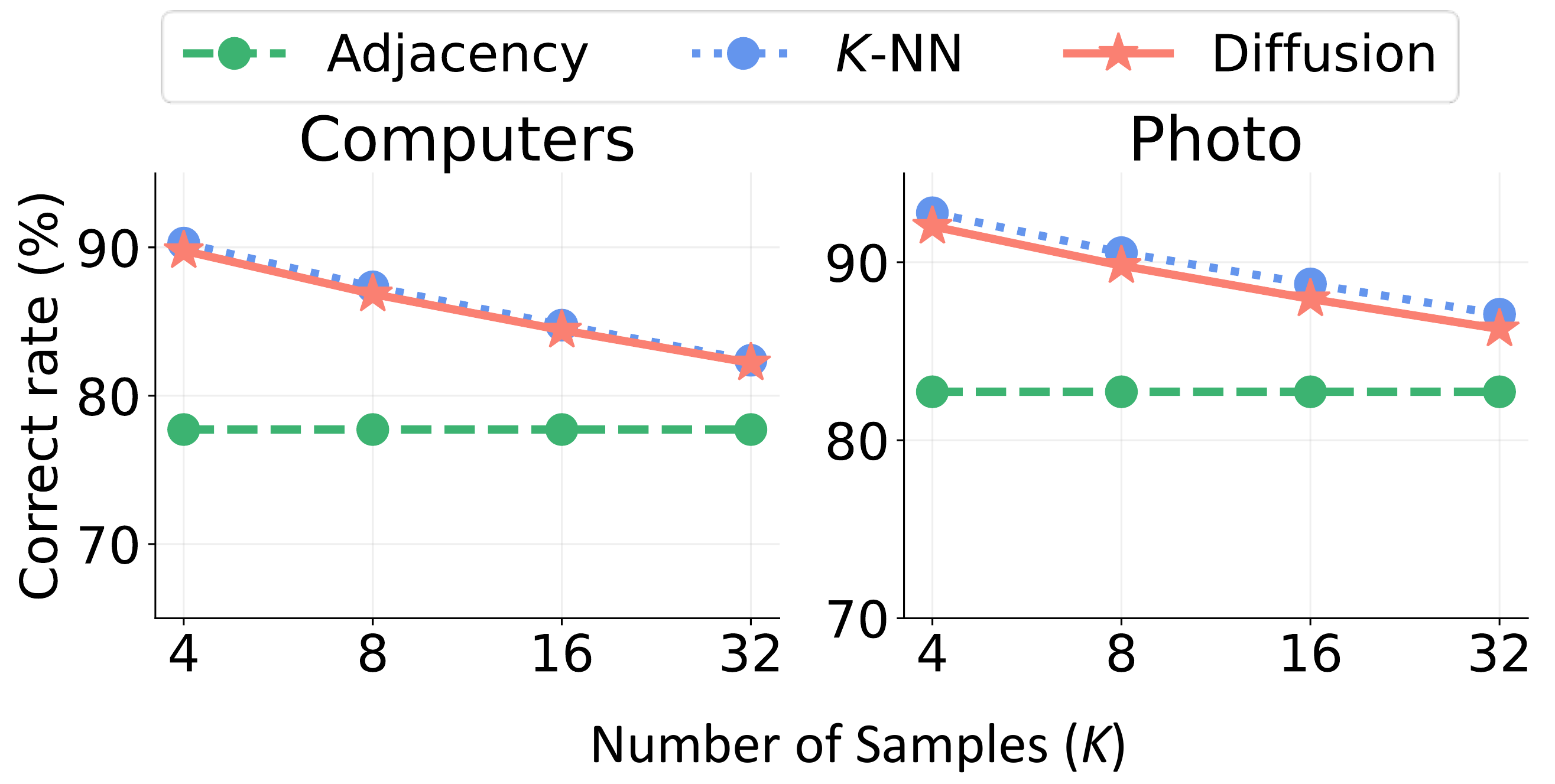} 
    \vspace{-2ex}
    \caption{Analysis on the ratio of its neighboring nodes being the same label as the query node across different $K$s.}
    \vspace{-3ex}
    \label{fig:diff_topk_adj}
\end{figure}

\textbf{1) Adjacency:} The most na\"ive approach is to consider the neighboring nodes of a query node as the set of local anchor nodes. Although the adjacency matrix contains local connectivity information among nodes, not all the adjacent relationship reveals identical semantics among nodes. As shown in Fig.~\ref{fig:diff_topk_adj}, the pure adjacency information contains many false positive relationships among the nodes, and thus it is not appropriate as $\mathbf{N}_i^{\text{local}}$. For example, more than 20\% and 15\% of neighboring nodes of a query node have different labels in Computers and Photo datasets on average, respectively. 
\textbf{2) $K$-NN:} Instead of relying on the adjacency information, we can use $K$-NN approach to sample $K$ nodes that are most similar to the query node according to the learned node representations. Although we can greatly reduce the number of false positives as shown in Fig.~\ref{fig:diff_topk_adj}, $K$-NN not only discovers nodes that are explicitly connected to the query node but also those that share similar features, i.e., distant but similar nodes, which contradicts with our intention of capturing the similarity among nodes in the local perspective. Moreover, it requires additional $\mathcal{O}(|\mathcal{V}|^2)$ space and time complexity, which makes it impractical to apply \proposed~on real-world large graphs. 
\textbf{3) Diffusion:} To this end, diffusion-based approach~\cite{klicpera2019diffusion} can be considered as a compromise between the above approaches. Graph diffusion calculates the closeness of nodes in the graph structure by repeatedly passing the weighting coefficients to the neighboring nodes. As shown in Fig.~\ref{fig:diff_topk_adj}, anchor nodes sampled with high diffusion scores have high probability of sharing the same label as much as $K$-NN even though diffusion only considers the structural information. Furthermore, diffusion does not require any additional computation during training since diffusion scores can be readily computed before the model training. Hence, in this paper, we leverage diffusion to sample local anchor nodes, i.e., $\mathbf{N}_i^{\text{local}}$.

More precisely, we calculate a diffusion matrix~\cite{klicpera2019diffusion} $\mathbf{S}$, based on personalized PageRank (PPR) \cite{page1999pagerank} as follows:
\begin{eqnarray}
\small
	\mathbf{S} = \sum_{k=0}^{\infty}{t(1-t)^{k}\mathbf{T}^{k}}
\end{eqnarray}
where $t \in (0, 1)$ is the teleport probability, and $\mathbf{T}$ is the symmetric transition matrix $\mathbf{T} = \mathbf{D}^{-1/2}\mathbf{A}\mathbf{D}^{-1/2}$, where $\mathbf{D}$ is the diagonal matrix of node degrees, i.e, $\mathbf{D}_{ii} = \sum_{j=1}^{N}{\mathbf{A}_{ij}}$. Each component $(i, j)$ of the diffusion matrix $\mathbf{S}$ indicates the closeness of node $v_i$ and $v_j$. 
For a query node $v_i$, we pre-define the top-$K$ highest scoring nodes denoted by $\mathbf{N}_i^{local}$ before the training of~\proposed.
During the training, we calculate the KL divergence loss defined in Eqn.~\ref{eqn:kld}, i.e., $\mathcal{L}_{\theta, \xi}^{\text{local}}$, between the similarity distributions $p^{\theta}_{\text{local}}$ and $p^{\xi}_{\text{local}}$ computed based on $\mathbf{N}_i^{\text{local}}$ through Eqn.~\ref{eqn:sim1} and Eqn.~\ref{eqn:sim2}, respectively. 

\subsection{Model update}
\label{sec:model update}
\subsubsection{\textbf{Updating online encoder $f_{\theta}$ and predictor $g_{\theta}$}}
During the training, the online parameters $\theta$ are updated to jointly minimize both global and local losses, i.e., $\mathcal{L}_{\theta, \xi}^{\text{glob}}$ and $\mathcal{L}_{\theta, \xi}^{\text{local}}$, as follows: $\mathcal{L}_{\theta, \xi} = \mathcal{L}_{\theta, \xi}^{\text{glob}} + \lambda \cdot \mathcal{L}_{\theta, \xi}^{\text{local}},\,\,\,\, \theta \leftarrow \operatorname{optimizer}\left(\theta, \nabla_{\theta} \mathcal{L}_{\theta, \xi}, \eta\right)$,
where $\lambda$ controls the importance of local structural information, and $\eta$ is the learning rate for online network. 

\subsubsection{\textbf{Updating target encoder $f_{\xi}$}}
\proposed~ updates the target encoder by smoothing the parameter of online encoder with the decay rate $\gamma$: $\xi\leftarrow\gamma\xi+(1-\gamma)\theta$.

\subsubsection{\textbf{Graph augmentation functions}}
Following previous work \cite{GRACE,BGRL}, we use standard graph perturbations for augmentation functions $\mathcal{T}_{1}$ and $\mathcal{T}_{2}$.
Specifically, we randomly mask node features and drop edges with fixed probabilities ($p_{f,1}$ and $p_{e,1}$ for $\mathcal{T}_{1}$, and $p_{f,2}$ and $p_{e,2}$ for $\mathcal{T}_{2}$).
We consider only simple and standard augmentation functions to study the effect of \proposed~as a representation learning framework.

\subsection{Extension to Heterogeneous Graphs}
\label{sec:multiplex}
Thanks to the generality of~\proposed~as a relational learning framework, it can be naturally extended to an attributed multiplex network~\cite{HDMI,DMGI,HAN}, which is a type of heterogeneous graph~\cite{shi2016survey}. 
As a multiplex network consists of multiple layers of attributed graphs, we consider each layer of the multiplex network as a view instead of generating multiple views through augmentations. 
Then, we make pairs with the layers to perform~\proposed. For example, given four layers of attributed graphs, we make six pairs of graphs in total, i.e., $4 \choose 2$ $=6$.
At inference time, we employ a mean pooling function to aggregate layer-specific node representations.

\section{Discussion}
\label{sec:Discussion}
In this section, we address limitations of previous state-of-the-art methods of two different types, i.e., contrastive learning (GRACE~\cite{GRACE} /GCA~\cite{GCA}) and non-contrastive learning (BGRL~\cite{BGRL}), and discuss how~\proposed~overcomes their limitations by leveraging relational information among nodes. 

\smallskip
\noindent\textbf{Comparison to contrastive learning-based methods.}
Under the principle of instance discrimination \cite{IND}, GRACE/GCA learns node representations by contrasting the representation of a query node with the representations of all other nodes in the two augmented graphs, while matching the representations of the same node.
That is, GRACE/GCA simply treats all other nodes apart from the query node itself as negatives even if they may share similar semantics with the query node, i.e., sampling bias~\cite{chuang2020debiased,lin2021prototypical}.
What's even worse here is that the softmax-based contrastive loss used in GRACE/GCA is a \textit{hardness-aware loss} function \cite{wang2021understanding}, which gives larger penalties to the anchor nodes that are closer (or more similar) to the query node.
Such a hardness-aware loss would facilitate the model to learn more discriminative boundary in the supervised setting where negative samples can be readily obtained.
However, this is not the case in self-supervised learning methods. 
As shown in Fig~\ref{fig:diff_topk_adj}, a large proportion of nearest-neighbors of a query node obtained by $K$-NN approach shares the same label as the query node. 
As all other nodes apart from the query node are treated as negatives, these close anchor nodes that share the same label as query node will not only be treated as negatives, but also be given more penalties as they are close to the query node, i.e., these false negative nodes will be trained to be even more dissimilar to the query node compared with other true negative nodes~\cite{wang2021understanding}. 
These false supervisory signals would seriously interfere with the representation learning process. 
On the other hand,~\proposed~relaxes the strict binary classification of GRACE/GCA with soft labeling so that the model can decide how much to push or pull other nodes based on the relational information among the nodes without relying on the binary decisions of positives and negatives.

\smallskip
\noindent\textbf{Comparison to non-contrastive learning-based method.}
BGRL learns node representations by constraining the representations from two encoders
to be close to each other without any use of negative samples, which addresses the aforementioned limitations of existing contrastive learning-based methods. 
However, as will be later shown in Fig.~\ref{fig:feature ablation}, we observe that BGRL suffers from a severe performance degradation when the feature information is not fully informative, i.e., contains noise, which is very common in reality. 
We attribute such behavior of BGRL to the strict self-preserving loss, {which considers each node to be independent from other nodes after message passing, and enforces each node to be augmentation-invariant.}
Due to the strict constraint, BGRL may overfit to a few non-informative features leading to a severe performance degradation when the features are noisy. However, we argue that the overfitting problem can be overcome with a little help from other nodes in the graph, i.e., learning from the relationship with other nodes.
To this end,~\proposed~relaxes the strict self-preserving loss with relation-preserving loss, allowing the representations to vary as long as the relationship among the representations is preserved. 
Thanks to the flexibility of the relation-preserving loss, \proposed~is robust to the quality of node features as will be shown in Fig. \ref{fig:feature ablation}.

\begin{table}[t]
    \centering
    \footnotesize
    \caption{Comparison on computational complexity}
    \vspace{-2ex}
    \begin{tabular}{c|c}
    Model   & Complexity \\ \hline\hline
    GRACE   & $4C_{\text{encoder}}(M+N) + 4C_{\text{projection}}N + C_{\text{GRACE}}(N^2)$ \\
    BGRL    & $6C_{\text{encoder}}(M+N) + 4C_{\text{prediction}}N + C_{\text{BGRL}}(N)$ \\ \hline
    \proposed    & $6C_{\text{encoder}}(M+N) + 4C_{\text{prediction}}N + C_{\text{\proposed}}(NK)$ \\ \hline
    \end{tabular}
    \label{tab:computational complexity}
    \vspace{-5ex}
\end{table}

\smallskip
\noindent\textbf{Computational Complexity Analysis.}
Assume we are given a graph with $N$ nodes and $M$ edges, and encoder $f$ that computes embeddings in time and space complexity of $\mathcal{O}(N + M)$. \proposed~and BGRL \cite{BGRL} perform four encoder computations per update step with an additional node-level prediction step, while GRACE \cite{GRACE} performs two encoder computations with a node-level projection step. 
Assuming that a backpropagation is approximately as costly as a forward pass, the total time and space complexities per update step are given in Table \ref{tab:computational complexity}, where $C$ are constants depending on the architecture of the different components, and $K$ is the number of samples in~\proposed, i.e., the size of the set of anchor nodes $\mathbf{N}_i$ for each node $v_i$.
As shown in Table \ref{tab:computational complexity}, the complexity of~\proposed~increases linearly with $K$. 
However, as shown in Section~\ref{sec:sensitivity},~\proposed~is robust over different $K$s, and thus we set $K$ to a value that is far smaller than the number of nodes $N$.
In summary, since $K \ll N$,~\proposed~is more efficient than GRACE, and only entails slight increase of complexity compared with BGRL.
Note that the node sampling probability and the diffusion matrix computation can be readily done before the model training, thus do not require any additional computational cost during the training process.

\section{Experiments}
\subsection{Experimental Setup}

\noindent\textbf{Datasets.}
We use \textbf{fourteen} widely used datasets to comprehensively evaluate the performance of~\proposed~on various downstream tasks, i.e., node classification and link prediction. The datasets include Wiki-CS~\cite{WikiCS}, Amazon (\textit{Computers} and \textit{Photo})~\cite{Amazon}, Coauthor (\textit{Co.CS} and \textit{Co.Physics})~\cite{MS}, Plantoid (\textit{Cora}, \textit{Citeseer} and \textit{Pubmed}), Cora Full~\cite{bojchevski2017deep}, ogbn-arXiv~\cite{hu2020open}, Reddit~\cite{Graphsage}, and protein-protein interaction network (\textit{PPI})~\cite{zitnik2017predicting,Graphsage}.
Moreover, we use IMDB and DBLP~\cite{DMGI} to evaluate \proposed~on node classification on multiplex networks.
The detailed statistics are summarized in Table \ref{tab:data_stats}.

\smallskip
\noindent\textbf{Methods Compared.}
We compare \proposed~with recent state-of-the-art graph representation learning methods, i.e., GRACE~\cite{GRACE}, GCA~\cite{GCA}, CCA-SSG~\cite{zhang2021canonical}, and BGRL~\cite{BGRL}. During the experiment, we use the official codes published by authors and then conduct evaluations within the same environment.
For node classification, we also report previously published results of raw features (Feats.) and other representative methods, such as node2vec (n2v) \cite{node2vec}, DeepWalk (DW) \cite{deepwalk}, DGI \cite{DGI}, GMI \cite{GMI}, and MVGRL \cite{MVGRL} as done in \cite{BGRL,GCA}.
For evaluations on multiplex networks, we compare \proposed~against recent state-of-the-art multiplex network representation learning methods, i.e., HAN \cite{HAN}, DMGI \cite{DMGI}, and HDMI \cite{HDMI}.

\begin{table}[t]
    \centering
    \footnotesize
    \caption{Statistics for datasets used for experiments.}
    \vspace{-2ex}
    \renewcommand{\arraystretch}{0.7}
    \resizebox{0.99\linewidth}{!}{
    \begin{tabular}{c|ccccc}
                        Dataset  & Type        & \# Nodes & \# Edges & \# Features & \# Cls. \\ \hline \hline
    WikiCS \tablefootnote{ \url{https://github.com/pmernyei/wiki-cs-dataset/raw/master/dataset}}                & reference & 11,701 & 216,123 & 300 & 10 \\
    Amazon Computers \tablefootnote{\label{url:shchur} \url{https://github.com/shchur/gnn-benchmark/raw/master/data/npz/}} & co-purchase & 13,752 & 245,861 & 767 & 10 \\
    Amazon Photo \textsuperscript{\ref{url:shchur}} & co-purchase & 7,650 & 119,081 & 745 & 8 \\
    Coauthor CS \textsuperscript{\ref{url:shchur}} & co-author   & 18,333 & 81,894 & 6,805 & 15 \\
    Coauthor Physics \textsuperscript{\ref{url:shchur}} & co-author   & 34,493 & 247,962 & 8,415 & 5 \\
    Cora \tablefootnote{\label{url:Planetoid} \url{https://github.com/kimiyoung/planetoid/raw/master/data}} & citation    & 2,708 & 5,429 & 1,433 & 7 \\
    Citeseer \textsuperscript{\ref{url:Planetoid}} & citation    & 3,327 & 4,732 & 3,703 & 6 \\
    Pubmed \textsuperscript{\ref{url:Planetoid}} & citation    & 19,717 & 44,338 & 500 & 3 \\
    Cora Full \tablefootnote{\url{https://github.com/abojchevski/graph2gauss/raw/master/data/cora.npz}} & citation & 19,793 & 65,311 & 8,710 & 70 \\
    ogbn-arXiv \tablefootnote{ \url{http://snap.stanford.edu/ogb/data/nodeproppred/arxiv.zip}} & citation    & 169,343 & 1,166,243 & 128 & 40 \\ \hline
    Reddit \tablefootnote{\url{https://data.dgl.ai/dataset/reddit.zip}} & community   & 231,443 & 11,606,919 & 602 & 41 \\
    PPI (24 Graphs) \tablefootnote{\url{https://data.dgl.ai/dataset/ppi.zip}} & interaction & 56,944 & 818,716 & 50 & 121 \\ \hline
    \multirow{2}{*}{IMDB \tablefootnote{\label{url:DMGI}\url{https://www.dropbox.com/s/48oe7shjq0ih151/data.tar.gz?dl=0}}} & co-actor    & \multirow{2}{*}{3,550} & 66,428    & \multirow{2}{*}{2,000} & \multirow{2}{*}{3} \\
                          & co-director &  & 13,788 &  &  \\
    \multirow{3}{*}{DBLP \textsuperscript{\ref{url:DMGI}}} & co-author   & \multirow{3}{*}{7,907} & 144,738   & \multirow{3}{*}{2,000} & \multirow{3}{*}{4} \\ 
                          & co-paper &  & 90,145 &  &  \\
                          & co-term &  & 57,137,515 &  &  \\ \hline
    \end{tabular}}
    \vspace{-4ex}
    \label{tab:data_stats}
\end{table}

\smallskip
\noindent\textbf{Evaluation Protocol.}
For node classification, we first train models in an unsupervised manner, and use the learned node representations to train and test a simple logistic regression classifier~\cite{DGI}. We use a random split of the nodes into train/validation/test nodes of 10/10/80\%, respectively, except for WikiCS, ogbn-arXiv, Reddit and PPI datasets for which public splits are given.
For link prediction, we first randomly split the original graph into train/validation/test edges of 50/20/30\%, i.e., $\text{E}_{\text{train}}$, $\text{E}_{\text{val}}$, and $\text{E}_{\text{test}}$, and generate the same amount of negative edges, i.e., $\text{E}_{\text{train}_{\text{neg}}}$, $\text{E}_{\text{val}_{\text{neg}}}$, and $\text{E}_{\text{test}_{\text{neg}}}$, from the nodes that are not connected in the original graph~\cite{SEAL}. 
We conduct experiments on two types of negative samples, i.e., random negatives that are sampled randomly among the pairs that are not directly connected in the original graphs, and hard negatives that are sampled among the pairs that are not directly connected but are located within three hop distances from the target node. 
After learning the node representations with $\text{E}_{\text{train}}$, we train a logistic regression classifier with $\text{E}_{\text{train}}$ and $\text{E}_{\text{train}_\text{neg}}$, and test on $\text{E}_{\text{val}}$, $\text{E}_{\text{val}_\text{neg}}$, $\text{E}_{\text{test}}$, and $\text{E}_{\text{test}_\text{neg}}$.
For both tasks, we report the test performance when the performance on validation set gives the best result.
We measure performance in terms of Accuracy, Macro-f1 and Micro-f1 for node classification and Area Under Curve (AUC) and Average Precision (AP) for link prediction.

\smallskip
\noindent\textbf{Implementation Details.}
Following previous works \cite{DGI,GRACE,BGRL}, we employ three distinct model architectures for various tasks (i.e. transductive learning, inductive learning on large graphs (Reddit dataset), and inductive learning on multiple graphs (PPI dataset)). We also follow the embedding dimensions, optimizer, activation function and augmentation hyperparameters of the state-of-the-art baseline, i.e., BGRL~\cite{BGRL}, while hyperparameters that are newly introduced for~\proposed~are tuned in certain ranges. 

\subsection{Performance Analysis}

\begin{table}[t]{
    \centering
    \footnotesize
    \caption{Performance on node classification tasks (OOM: Out of Memory on 24GB RTX3090).}
    \vspace{-3ex}
    \renewcommand{\arraystretch}{0.99}
    \resizebox{0.99\linewidth}{!}{
    \begin{tabular}{p{1cm}|p{1.12cm}p{1.12cm}p{1.12cm}p{1.12cm}p{1.12cm}}
        & WikiCS & Computers & Photo  & Co.CS       & Co.Physics  \\ \hline \hline
        GCN        & 77.19 \scriptsize{(0.12)} & 86.51 \scriptsize{(0.54)} & 92.42 \scriptsize{(0.22)} & 93.03 \scriptsize{(0.31)} & 95.65 \scriptsize{(0.16)} \\ \hline 
        Feats.        & 71.98 \scriptsize{(0.00)} & 73.81 \scriptsize{(0.00)} & 78.53 \scriptsize{(0.00)} & 90.37 \scriptsize{(0.00)} & 93.58 \scriptsize{(0.00)} \\
        n2v            & 71.79 \scriptsize{(0.05)} & 84.39 \scriptsize{(0.08)} & 89.67 \scriptsize{(0.12)} & 85.08 \scriptsize{(0.03)} & 91.19 \scriptsize{(0.04)} \\
        DW            & 74.35 \scriptsize{(0.06)} & 85.68 \scriptsize{(0.06)} & 89.44 \scriptsize{(0.11)} & 84.61 \scriptsize{(0.22)} & 91.77 \scriptsize{(0.15)} \\
        DW+Feats. & 77.21 \scriptsize{(0.03)}  & 86.28 \scriptsize{(0.07)} & 90.05 \scriptsize{(0.08)} & 87.70 \scriptsize{(0.04)} & 94.90 \scriptsize{(0.09)} \\ \hline
        DGI                 & 75.35 \scriptsize{(0.14)} & 83.95 \scriptsize{(0.47)} & 91.61 \scriptsize{(0.22)} & 92.15 \scriptsize{(0.63)} & 94.51 \scriptsize{(0.52)} \\
        GMI                 & 74.85 \scriptsize{(0.08)} & 82.21 \scriptsize{(0.31)} & 90.68 \scriptsize{(0.17)} & OOM & OOM \\
        MVGRL               & 77.52 \scriptsize{(0.08)} & 87.52 \scriptsize{(0.11)} & 91.74 \scriptsize{(0.07)} & 92.11 \scriptsize{(0.12)} & 95.33 \scriptsize{(0.03)} \\
        GRACE         & 78.25 \scriptsize{(0.65)} & 88.15 \scriptsize{(0.43)} & 92.52 \scriptsize{(0.32)} & 92.60 \scriptsize{(0.11)} & OOM \\
        GCA           & 78.30 \scriptsize{(0.62)} & 88.49 \scriptsize{(0.51)} & 92.99 \scriptsize{(0.27)} & 92.76 \scriptsize{(0.16)} & OOM \\ 
        CCA-SSG         & 77.88 \scriptsize{(0.41)} & 87.01 \scriptsize{(0.41)} & 92.59 \scriptsize{(0.25)} & 92.77 \scriptsize{(0.17)} & 95.16 \scriptsize{(0.10)} \\ 
        BGRL          & 79.60 \scriptsize{(0.60)} & 89.23 \scriptsize{(0.34)} & 93.06 \scriptsize{(0.30)} & 92.90 \scriptsize{(0.15)} & 95.43 \scriptsize{(0.09)} \\ \hline
        \proposed    & \textbf{80.29 \scriptsize{(0.72)}} & \textbf{89.70 \scriptsize{(0.44)}} & \textbf{93.43 \scriptsize{(0.31)}} & \textbf{92.94 \scriptsize{(0.13)}} & \textbf{95.46 \scriptsize{(0.10)}}\\ \hline
    \end{tabular}}
    \vspace{-2ex}
    \label{tab:main results}
    }
\end{table}

\begin{figure}[t]{
    \centering
    \includegraphics[width=0.7\columnwidth]{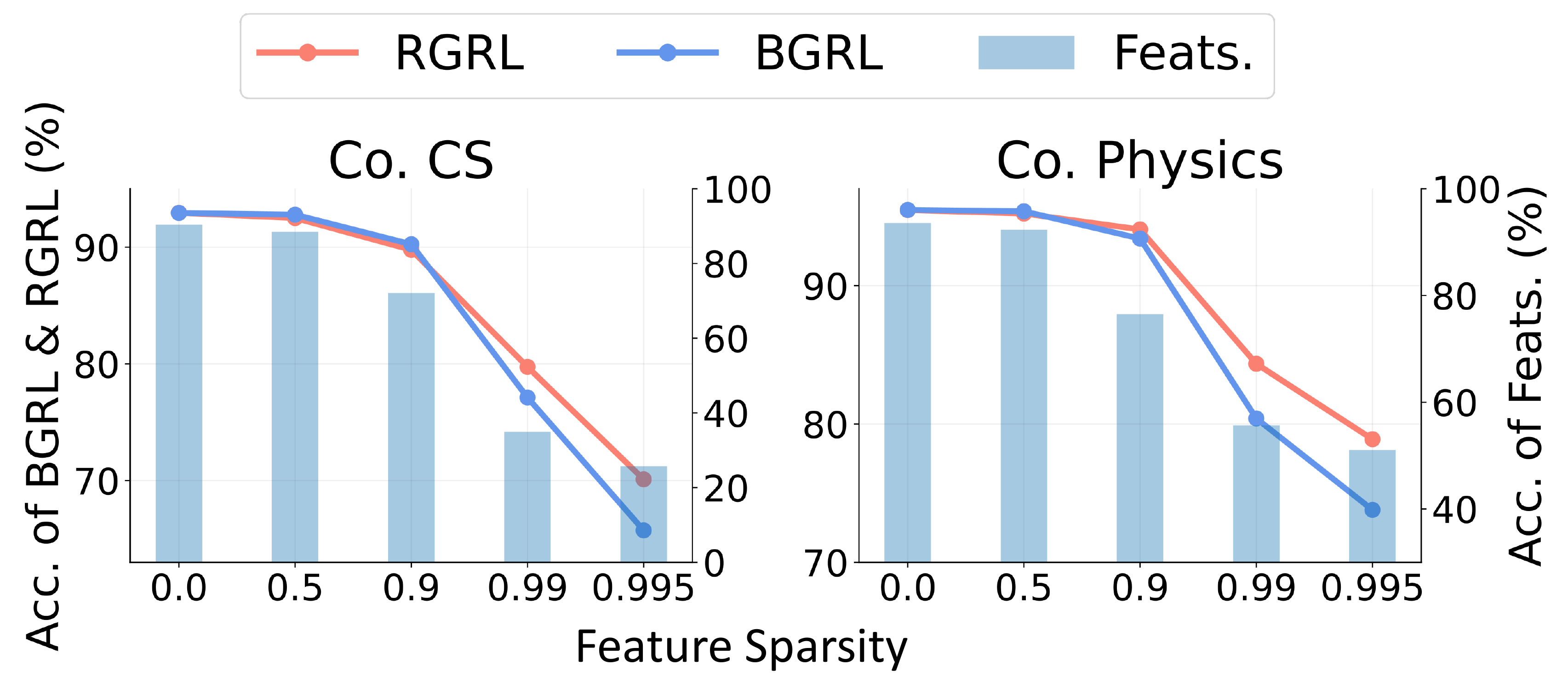} 
    \vspace{-2ex}
    \caption{Node classification accuracy over feature sparsity.~\proposed~is robust to low-quality features.
    }
    \label{fig:feature ablation}
    \vspace{-3ex}
}\end{figure}

\noindent\textbf{Evaluation on node classification.}
The empirical performance on node classification is summarized in Table \ref{tab:main results}. We have the following observations: 
\textbf{1)} Our relational learning framework \proposed~ outperforms all other baseline methods that overlook the relationship among nodes, i.e., GRACE, GCA, CCA-SSG, and BGRL. We argue that as \proposed~allows the node representations to vary as long as the relationship with other nodes is preserved, the learned node representations capture the relational information among the nodes, thereby improving the performance of node classification. Considering that graphs reveal relationship among nodes, our relational learning framework aligns with the very nature of graph-structured data, which have been overlooked in previous methods.
\textbf{2)} It is worth noting that methods built upon the instance discrimination principle~\cite{IND}, i.e., GRACE and GCA, not only are memory consuming (OOM on large datasets), but also generally perform worse than their counterpart, i.e., BGRL. This indicates that instance discrimination, which treats all other nodes except itself as negatives without considering the graph structural information, is not appropriate for graph-structured data that contain relational information among nodes.
\textbf{3)} We find out that ~\proposed~shows consistent improvements in datasets whose given feature is less informative (i.e., WikiCS, Computers, and Photo) as shown by the performance of ``Feats.'' in Table \ref{tab:main results}. Thanks to the external self-supervisory signals from other nodes, \proposed~performs well even without rich information of a single node. 
To corroborate the results, we conduct additional experiments by randomly corrupting a certain ratio of the input features of Co.CS and Co.Physics datasets whose features are relatively more informative as shown in Fig.~\ref{fig:feature ablation}.
We indeed observe that \proposed~is more robust than BGRL as the quality of the input features gets worse. We argue that this is mainly because~\proposed~learns the representation of a node from its relationship with other nodes in the graph rather than relying on the information contained in the node itself.

We also conduct experiments on commonly used small datasets (i.e., {Cora}, {Citeseer}, {Pubmed}, {CoraFull}), and a large dataset (i.e., {ogbn-arXiv}) as shown in Table \ref{tab:additional results}. For ogbn-arXiv, we report results on both validation and test sets following \cite{BGRL}, since the dataset is split according to the chronological order.
\textbf{1)} As shown in Table \ref{tab:additional results}, \proposed~outperforms all other recent self-supervised learning methods on various datasets.
\textbf{2)} Lastly, we conduct inductive node classification on Reddit and PPI datasets. We observe that~\proposed~performs the best, which demonstrates the inductive capability of~\proposed. 
We attribute this to the flexibility of the relation-preserving loss of~\proposed~that helps avoid overfitting to the training data compared with the previous methods as discussed in Sec.~\ref{sec:Discussion}, which eventually leads to improvements on unseen data.


\begin{table}[t]
    \centering
    \footnotesize
    \caption{Performance on transductive node classification on other datasets (Accuracy), and inductive node classification on Reddit and PPI datasets (Micro-F1).}
    \vspace{-3ex}
    \renewcommand{\arraystretch}{0.8}
    \resizebox{0.99\linewidth}{!}{
    \begin{tabular}{c|cccccc|cc}
     & \multicolumn{6}{c|}{Transductive} & \multicolumn{2}{c}{Inductive} \\ \cline{2-9} 
    & \multirow{2}{*}{Cora}      & Cite-                      & Pub-                       & \multicolumn{1}{l|}{Cora}    & \multicolumn{2}{c|}{ogbn-arXiv}                          & \multirow{2}{*}{Reddit}    & \multirow{2}{*}{PPI}       \\ \cline{6-7}
     &  & seer & med & \multicolumn{1}{l|}{Full} & Valid & Test &  & \\ \hline \hline
    \multirow{2}{*}{GRACE} & 83.38 & 70.79 & 83.96 & \multicolumn{1}{l|}{64.19} & \multirow{2}{*}{OOM} & \multirow{2}{*}{OOM} & 94.84 & 67.12 \\
    & \multicolumn{1}{c}{\scriptsize{(0.95)}} & \multicolumn{1}{c}{\scriptsize{(0.83)}} & \multicolumn{1}{c}{\scriptsize{(0.29)}} & \multicolumn{1}{c|}{\scriptsize{(0.36)}} & & & \multicolumn{1}{c}{\scriptsize{(0.03)}} & \multicolumn{1}{c}{\scriptsize{(0.05)}} \\
    \multirow{2}{*}{GCA} & 82.79 & 70.70 & 84.19 & \multicolumn{1}{l|}{64.34} & \multirow{2}{*}{OOM} & \multirow{2}{*}{OOM} & 94.85 & 66.72 \\
    & \multicolumn{1}{c}{\scriptsize{(1.01)}} & \multicolumn{1}{c}{\scriptsize{(0.91)}} & \multicolumn{1}{c}{\scriptsize{(0.32)}} & \multicolumn{1}{c|}{\scriptsize{(0.42)}}  & & & \multicolumn{1}{c}{\scriptsize{(0.06)}} & \multicolumn{1}{c}{\scriptsize{(0.08)}} \\
    \multirow{2}{*}{CCA-SSG} & 83.01 & 70.35 & 84.81 & \multicolumn{1}{l|}{64.09} & 59.43 & 58.50 & 94.89 & 66.09 \\
    & \multicolumn{1}{c}{\scriptsize{(0.66)}} & \multicolumn{1}{c}{\scriptsize{(1.23)}} & \multicolumn{1}{c}{\scriptsize{(0.22)}} & \multicolumn{1}{c|}{\scriptsize{(0.37)}}  & \multicolumn{1}{c}{\scriptsize{(0.05)}} & \multicolumn{1}{c|}{\scriptsize{(0.08)}} & \multicolumn{1}{c}{\scriptsize{(0.02)}} & \multicolumn{1}{c}{\scriptsize{(0.01)}} \\
    \multirow{2}{*}{BGRL} & 82.82 & 69.06 & \textbf{86.16} & \multicolumn{1}{l|}{63.94} & 70.66 & 69.61 & 94.90  & 68.89 \\
    & \scriptsize{(0.86)} & \scriptsize{(0.80)} & \textbf{\scriptsize{(0.19)}} & \multicolumn{1}{c|}{\scriptsize{(0.39)}} & \scriptsize{(0.06)} & \scriptsize{(0.09)} & \scriptsize{(0.04)} & \scriptsize{(0.08)} \\ \hline 
    \multirow{2}{*}{\proposed}    & \textbf{83.98}  & \textbf{71.29}  & 85.33  & \multicolumn{1}{l|}{\textbf{64.62}}   & \textbf{72.34}  & \textbf{71.49}  & \textbf{95.04}  & \textbf{69.28}  \\
    & \textbf{\scriptsize{(0.78)}} & \textbf{\scriptsize{(0.87)}} & \scriptsize{(0.20)} & \multicolumn{1}{c|}{\textbf{\scriptsize{(0.39)}}} & \textbf{\scriptsize{(0.09)}} & \textbf{\scriptsize{(0.08)}} & \textbf{\scriptsize{(0.03)}} & \textbf{\scriptsize{(0.06)}} \\ \hline
    \end{tabular}}
    \label{tab:additional results}
    \vspace{-2ex}
\end{table}

\begin{table}{
    \centering
    \footnotesize
    \caption{Performance on link prediction with random and hard negative edges.}
    \vspace{-2ex}
    \renewcommand{\arraystretch}{0.95}
    \resizebox{0.99\linewidth}{!}{
    \begin{tabular}{cc|cccccccc}
            & & \multicolumn{2}{c}{Computers} & \multicolumn{2}{c}{Photo}    & \multicolumn{2}{c}{Co. CS}   & \multicolumn{2}{c}{Co. Physics} \\ \cline{3-10}
            & & AUC  & \multicolumn{1}{c}{AP} & AUC & \multicolumn{1}{c}{AP} & AUC & \multicolumn{1}{c}{AP} & AUC             & AP            \\ \hline \hline
    \multirow{5}{*}{\rotatebox[origin=c]{90}{Random Neg.}} & \multicolumn{1}{|c|}{GRACE}    & 0.939 & 0.939 & 0.962 & 0.960 & 0.970 & 0.970 & OOM & OOM \\
    \multicolumn{1}{l}{} & \multicolumn{1}{|c|}{GCA}     & 0.954 & 0.954 & 0.965 & 0.960 & \textbf{0.971} & \textbf{0.970} & OOM & OOM \\
    \multicolumn{1}{l}{} & \multicolumn{1}{|c|}{CCA-SSG} & 0.961 & 0.959 & 0.973 & 0.970 & 0.949 & 0.950 & 0.943 & 0.936 \\
    \multicolumn{1}{l}{}& \multicolumn{1}{|c|}{BGRL}    & 0.964 & 0.961 & 0.978 & 0.976 & 0.952 & 0.948 & 0.952 & 0.947 \\ \cline{2-10}
    \multicolumn{1}{l}{} & \multicolumn{1}{|c|}{\proposed}    & \textbf{0.974} & \textbf{0.972} & \textbf{0.983} & \textbf{0.981} & 0.967 & 0.968 & \textbf{0.964} & \textbf{0.963} \\ \hline \hline
    \multirow{5}{*}{\rotatebox[origin=c]{90}{Hard Neg.}} & \multicolumn{1}{|c|}{GRACE}   & 0.933 & 0.933 & 0.939 & 0.929 & 0.870 & 0.868 & OOM & OOM \\
    \multicolumn{1}{l}{} & \multicolumn{1}{|c|}{GCA}     & 0.938 & 0.929 & 0.948 & 0.939 & 0.874 & 0.869 & OOM & OOM \\
    \multicolumn{1}{l}{} & \multicolumn{1}{|c|}{CCA-SSG} & 0.954 & 0.952 & 0.947 & 0.943 & 0.847 & 0.835 & 0.871 & 0.856 \\
    \multicolumn{1}{l}{} & \multicolumn{1}{|c|}{BGRL}    & 0.959 & 0.956 & 0.959 & 0.956 & 0.845 & 0.832 & 0.903 & 0.892 \\ \cline{2-10}
    \multicolumn{1}{l}{} & \multicolumn{1}{|c|}{\proposed}    & \textbf{0.969} & \textbf{0.968} & \textbf{0.967} & \textbf{0.964} & \textbf{0.878} & \textbf{0.881} & \textbf{0.923} & \textbf{0.919} \\ \hline
    \end{tabular}}
    \label{tab:link prediction}
}\end{table}

\smallskip
\noindent\textbf{Evaluation on link prediction.} Table \ref{tab:link prediction} shows the link prediction performance on random and hard negative edges.
We have the following observations:
\textbf{1)}~\proposed~discovers the true existing relationships between the nodes better than all other baseline methods. This implies that learning augmentation-invariant relationship among nodes is beneficial not only for node classification but also for link prediction.
\textbf{2)} The improvements of~\proposed~on hard negative edges, which is a more practical task, is more significant than that on random negative edges implying that~\proposed~detects more fine-grained relational information compared with other baselines. 
\textbf{3)} In the case of Computers and Photo datasets, non-contrastive methods (i.e., BGRL, CCA-SSG) outperform contrastive methods (i.e., GRACE, GCA), while this is not the case for Coauthor CS and Physics datasets. On the other hand,~\proposed~outperforms all other baseline methods regardless of the datasets, since \proposed~relaxes the strict constraints of both contrastive/non-contrastive methods thereby achieving the best of both worlds as discussed in Sec. \ref{sec:Discussion}.

\smallskip
\noindent\textbf{Evaluation on multiplex network.} Table \ref{tab:multiplex} shows the node classification performance on multiplex networks. \proposed~outperforms previous state-of-the-art methods for multiplex networks showing the generality of our relational learning framework. 
To demonstrate the generality of \proposed, we apply BGRL to multiplex networks closely following the extension of \proposed. 
We observe that~\proposed~outperforms all other baseline methods whereas BGRL shows competitive performance with baseline methods. We argue that this is mainly due to the strict constraint of BGRL in that BGRL learns layer-invariant features whereas \proposed~learns layer-invariant relationships as discussed in Sec.~\ref{sec:Discussion}. Since \proposed~allows the node representations to vary as long as the core relationship among the multiple layers is kept, we argue that~\proposed~can learn from more diverse relationship inherent in multiplex network.

\begin{table}[t]
    \centering
    \footnotesize
    \caption{Performance on multiplex network.}
    \vspace{-3ex}
    \begin{tabular}{c|cc|cc}
    Dataset   & \multicolumn{2}{c|}{IMDB} & \multicolumn{2}{c}{DBLP}                 \\ \hline
    Metric    & Macro-F1    & Micro-F1    & Macro-F1 & Micro-F1 \\ \hline \hline
    HAN       & 0.599 & 0.607 & 0.716 & 0.708 \\
    DMGI      & 0.648 & 0.648 & 0.771 & 0.766 \\
    $\text{DMGI}_{\text{attn}}$ & 0.602 & 0.606 & 0.778 & 0.770 \\
    HDMI      & 0.650 & \textbf{0.658} & 0.820 & 0.811 \\ \hline
    BGRL      & 0.631 & 0.634 & 0.819 & 0.807 \\
    \proposed      & \textbf{0.653} & \textbf{0.658} & \textbf{0.830} & \textbf{0.818} \\ \hline
    \end{tabular}
    \label{tab:multiplex}
\end{table}

\begin{figure}[t]{
    \centering
    \includegraphics[width=0.8\columnwidth]{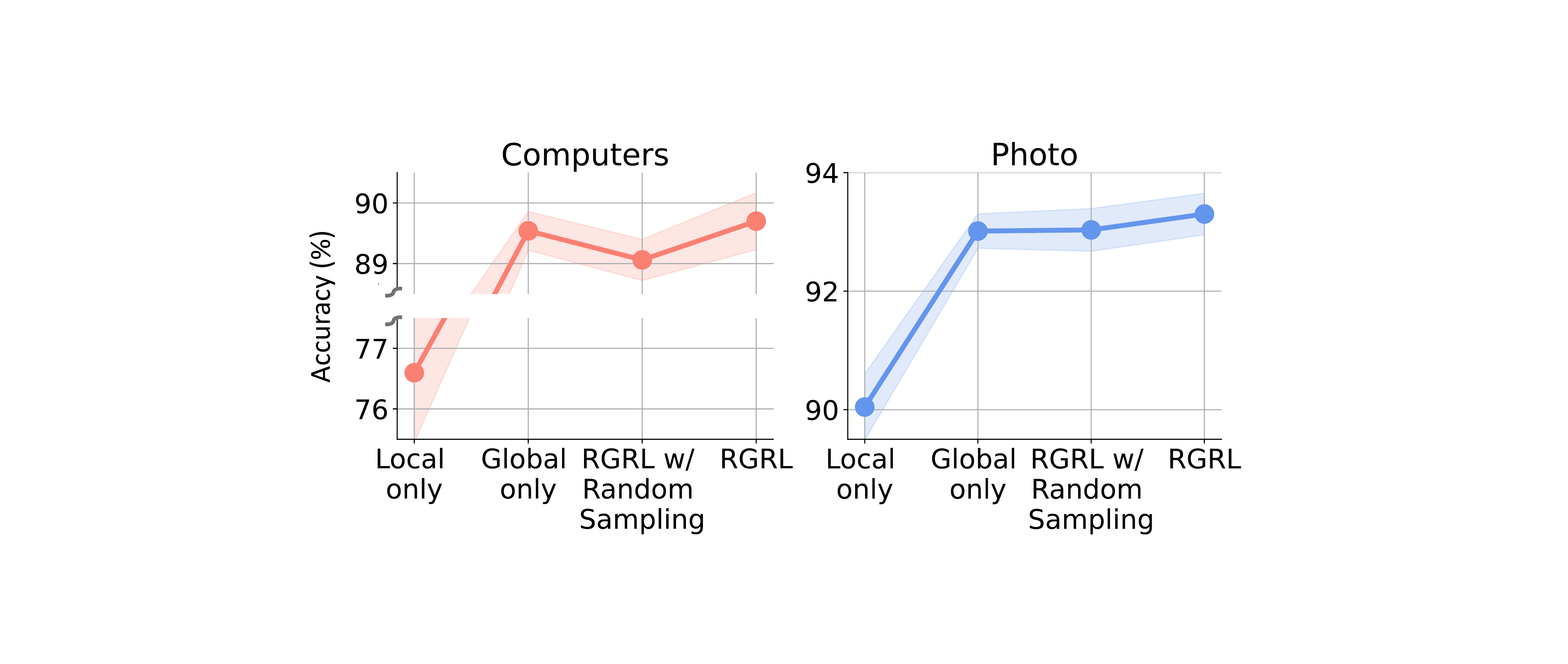} 
    \vspace{-2ex}
    \caption{Ablation studies.}
    \label{fig:ablation studies}
    \vspace{-2ex}
}\end{figure}

\subsection{Ablation Studies}

\begin{figure}[t]{
    \centering
    \includegraphics[width=0.99\columnwidth]{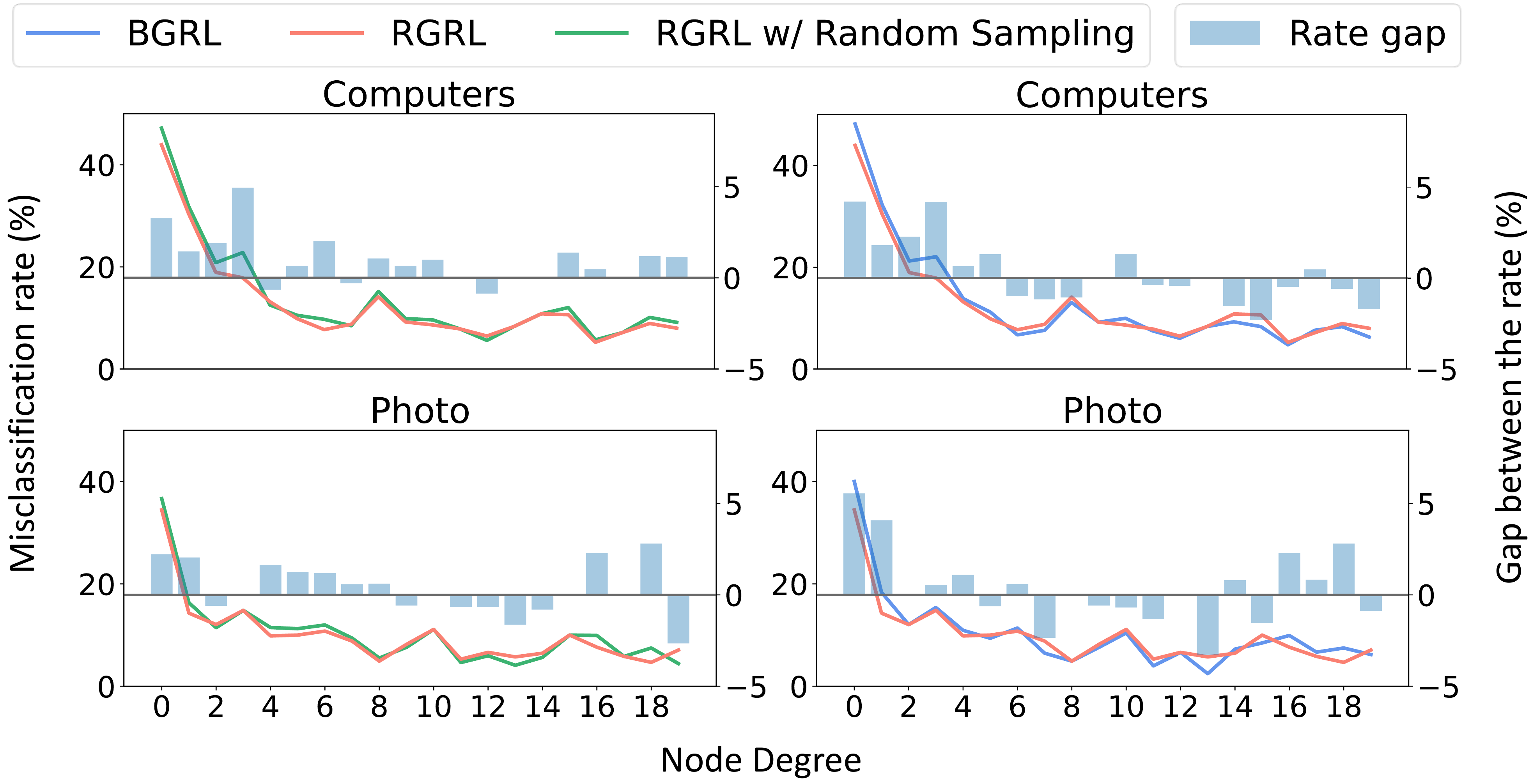}
    \vspace{-2ex}
    \caption{Comparisons of misclassification rate per node degree. (right) \proposed~vs.~\proposed~with random sampling (left)~\proposed~vs. BGRL.
    Rate gap indicates how well~\proposed~performs compared with the baseline. }
    \label{fig:wrong ratio per degree}
    }
\end{figure}

To verify the benefit of each component of \proposed, we conduct ablation studies on two datasets, i.e., Amazon Computers and Amazon Photo in Fig.~\ref{fig:ablation studies}. We have the following observations: 
\textbf{1)} Considering the global similarity is more beneficial than considering the local similarity. We argue that this is because the global similarity inherently reflects the similarity with both local and non-local nodes thanks to the anchor sampling process, whereas local similarity only reflects the local information. 
\textbf{2)} However, considering the local similarity in addition to the global similarity, i.e.,~\proposed, shows the best performance.
This is because the local similarity module captures the fine-grained relationship among structurally close nodes.
\textbf{3)} \proposed~outperforms ``\textit{\proposed~w/ Random Sampling}. This implies that~\proposed~successfully alleviates the degree-bias issue by sampling anchor nodes from the inverse degree-weighted distribution. To verify this, we plot the misclassification rate of~\proposed~per node degree, and compare it with~``\textit{\proposed~w/ Random Sampling} in Fig.~\ref{fig:wrong ratio per degree} (left). We observe that \proposed~outperforms ``\textit{\proposed~w/ Random Sampling} on low-degree nodes while competitive on other nodes, demonstrating that sampling anchor nodes from the inverse degree-weighted distribution helps alleviate degree-bias issue.
\textbf{4)} Moreover, we also compare the misclassification rate of~\proposed~per node degree with BGRL in Fig.~\ref{fig:wrong ratio per degree} (right). We observe that~\proposed~greatly outperforms BGRL especially on low-degree nodes, indicating the superiority of the relation-preserving loss of~\proposed~over the self-preserving loss of BGRL for alleviating the degree-bias issue of GNNs. As most real-world graphs are long-tailed, i.e., a majority of nodes have low degree, we argue that~\proposed~is practical for use in reality.

\begin{figure}[t]
    \centering
    \includegraphics[width=0.99\columnwidth]{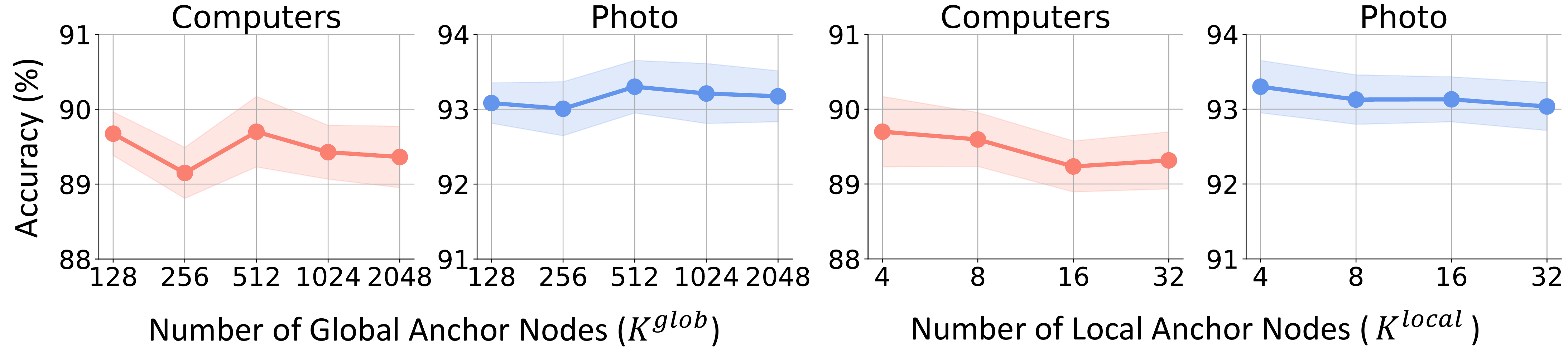} 
    \vspace{-2ex}
    \caption{Sensitivity analysis on number of global anchor nodes (Left) and local anchor nodes (Right).}
    \label{fig:sensitivity_sample}
    \vspace{-2ex}
\end{figure}

\subsection{Sensitivity Analysis}
\label{sec:sensitivity}

\begin{table*}
    \begin{minipage}{.37\linewidth}{
    \centering
    \includegraphics[width=0.98\linewidth]{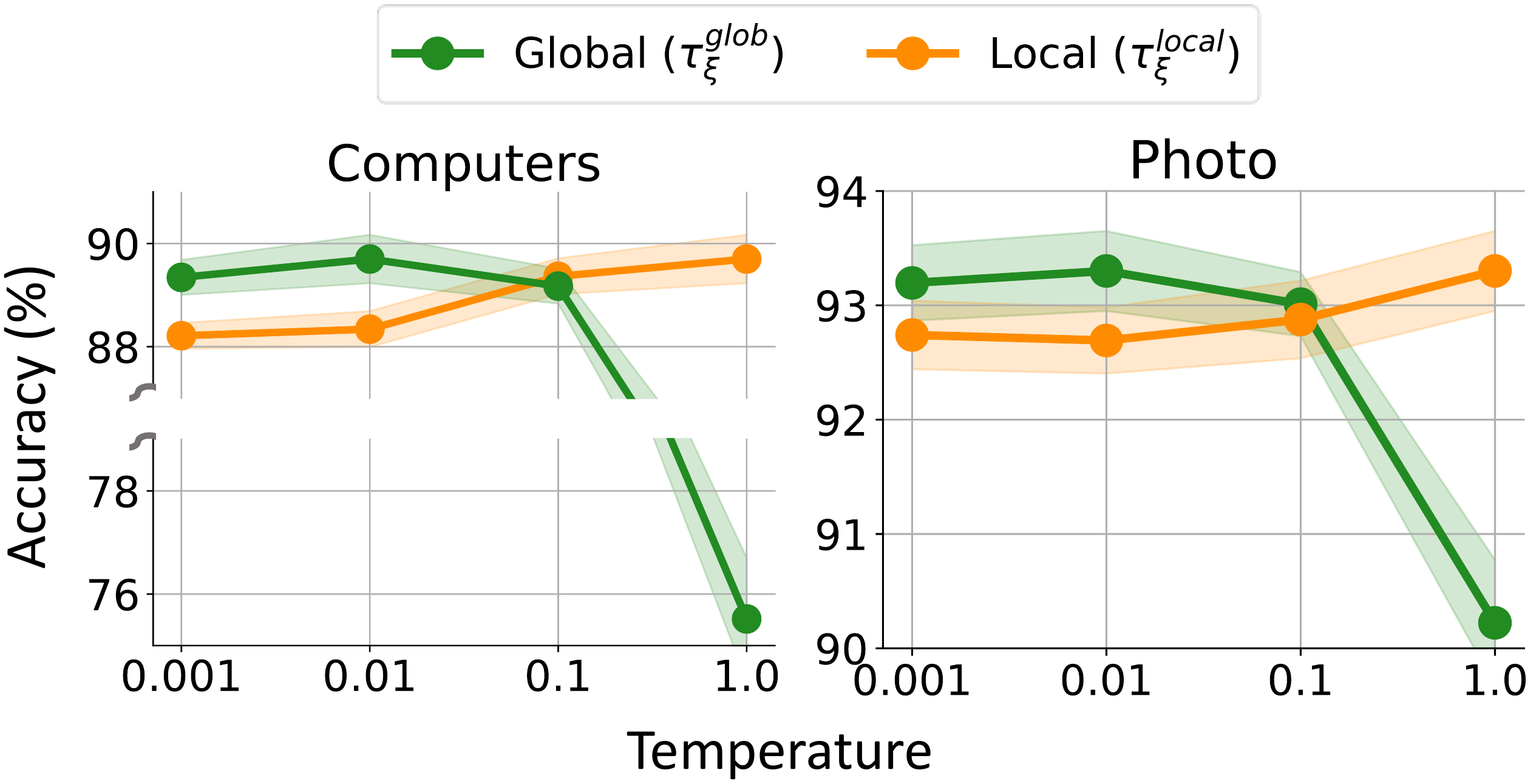} 
    \captionof{figure}{Sensitivity analysis on temperature.}
    \label{fig:sensitivity_temp}
    }
    \end{minipage}
    \begin{minipage}{0.58\linewidth}{
    \centering
    \includegraphics[width=0.92\linewidth]{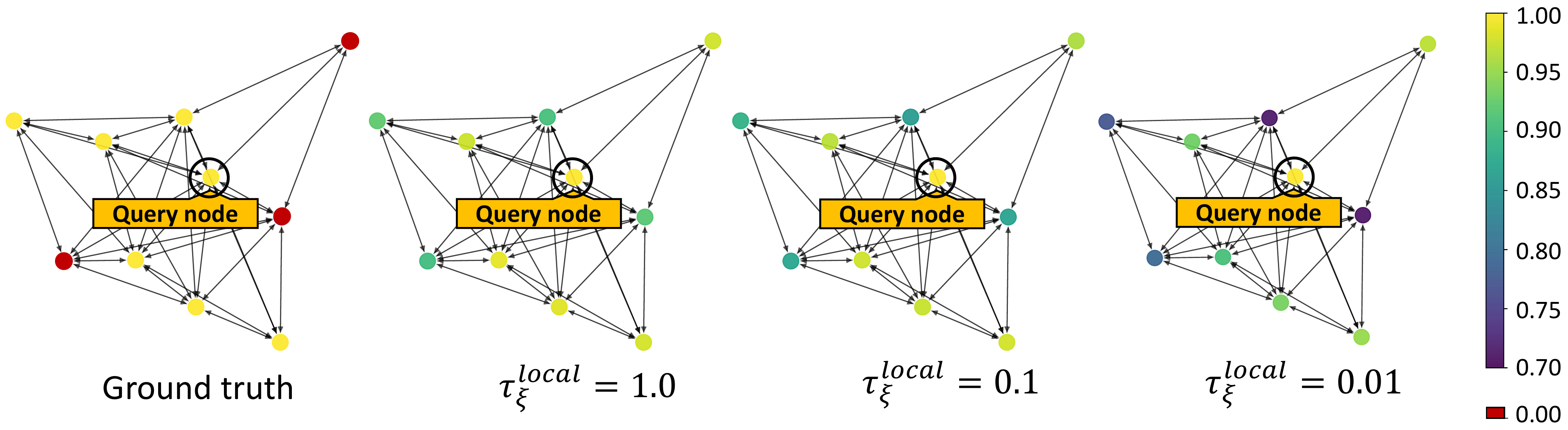} 
    \vspace{-3ex}
    \captionof{figure}{How $\tau_{\xi}^{\text{local}}$ reflects the structural information around the query node.  Color indicates the cosine similarity between the query node and its neighboring nodes. For Ground truth, the similarity is 1 if the neighboring node shares the same label with query node, and 0 in otherwise.}
    \label{fig:qualitative_temp}
    }\end{minipage}
\vspace{-1ex}
\end{table*}

\smallskip
\noindent\textbf{On the number of sampled anchor nodes.}
Fig.~\ref{fig:sensitivity_sample} shows the sensitivity analysis on the number of local and global anchor nodes, i.e., $K^{\text{local}}$ and $K^{\text{glob}}$.
We have the following observations:
\textbf{1)} Regarding the number of global anchor nodes, i.e., $K^{\text{glob}}$, we observe that the performance of node classification is relatively stable over various $K^{\text{glob}}$s. In other words, sampling more global anchor nodes does not contribute much to the model performance.
We argue that since different global anchor nodes are sampled in every training epoch, the computed similarity distributions would be diverse, which in turn facilitates the model to learn from diverse supervisory signals even with a small number of sampled nodes.
Thanks to its robustness to the number of global anchor nodes, \proposed~is practical for applying on large-scale graphs common in real-world, such as \textit{ogbn-arXiv} in Table \ref{tab:additional results}.
\textbf{2)} Regarding the number of local anchor nodes, i.e., $K^{\text{local}}$, we observe a slight performance degradation as we increase $K^{\text{local}}$. 
This is expected since more false positives are introduced to $\text{N}_{i}^{\text{local}}$ as $K^{\text{local}}$ gets larger as shown in Fig. \ref{fig:diff_topk_adj}.

\noindent\textbf{On the temperature hyperparameters.}
Fig.~\ref{fig:sensitivity_temp} shows the sensitivity analysis on the temperature hyperparameters $\tau_{\xi}^{\text{glob}}$ and $\tau_{\xi}^{\text{local}}$ of the target encoder. 
Note that temperature hyperparameters (i.e., $\tau_{\theta}^{\text{glob}}$ and $\tau_{\theta}^{\text{local}}$) of the online network are fixed to $0.1$. We have the following observations:
\textbf{1)} Regarding the global temperature hyperparameters,
we observe that the best $\tau_{\xi}^{\text{glob}}$ is 0.01 in both datasets, which is smaller than $\tau_{\theta}^{\text{glob}}$ that is fixed to 0.1.
Note that the target distribution gets sharper as $\tau_{\xi}^{\text{glob}}$ gets smaller. This aligns with the argument in~\cite{zheng2021ressl} that target distribution should be sharpened to provide a stronger supervisory signal for the model training.
Thus, with $\tau_{\xi}^{\text{glob}}$ smaller than $\tau_{\theta}^{\text{glob}}$, \proposed~can learn more discriminative node representations.
\textbf{2)} On the other hand, regarding the local temperature hyperparameters, we observe that a high value of $\tau_{\xi}^{\text{local}}$ is more beneficial for the model performance, i.e., $\tau_{\xi}^{\text{local}}=1.0$ shows the best performance in both datasets.
Since target distribution $p_{\xi}^{\text{local}}$ gets softened as $\tau_{\xi}^{\text{local}}$ gets larger, the model is trained to be less discriminative among the sampled local anchor nodes $\text{N}_{i}^{\text{local}}$.
That is, the model is trained to learn a set of local anchor nodes $\text{N}_{i}^{\text{local}}$, which are structurally close and share the same semantic, to be close in representation space.
As shown in Fig.~\ref{fig:qualitative_temp}, given a query node, the cosine similarities of its structurally close nodes are directly correlated with the choice of $\tau_{\xi}^{\text{local}}$; the similarity with the structurally close nodes increases as $\tau_{\xi}^{\text{local}}$ gets larger. This implies that~\proposed~can adaptively choose an appropriate $\tau_{\xi}^{\text{local}}$ regarding how much fine-grained relationship should be preserved in the representation space.

\vspace{-1ex}
\subsection{Qualitative Analysis}
We conduct a qualitative analysis to further demonstrate the effectiveness of \proposed~in capturing the relational information.
Our goal is to show the superiority of \proposed~ over BGRL at 1) discovering core relationships from the given graph, and 2) discovering meaningful knowledge that is not revealed in the given graph. For our analysis, we take ``Jiawei Han'' and ``Christos Faloutsos'' as the query authors, both of whom are renowned professors in the field of data mining.

\textbf{Case 1)} In Table \ref{tab:case1}, we present a case study that shows which author is the most similar based on the cosine similarity of the learned node representations. 
In both cases, we observe that the top-1 similar authors discovered by~\proposed~indeed have more co-authored papers\footnote{Since Co.CS dataset only provides the co-author relationship among authors, but not how many papers they co-authored, we tried our best to count the number of co-authored papers based on ``Google Scholar'' along with other resources such as authors' websites.} with the query authors compared with BGRL.
Surprisingly,
the top-1 similar authors discovered by~\proposed~happen to be former Ph.D. students of the query authors. 
Considering that the advisor-advisee relationship is one of the core relationships in the academia network that should be preserved no matter how the graph is perturbed, we argue that the relation-preserving framework of~\proposed~is effective.
\textbf{Case 2)} Table \ref{tab:case2} shows the top-1 similar authors among the authors who do not have any co-authored papers with the query authors. In the case of ``Jiawei Han'', the top-1 similar author discovered by BGRL is ``Zhou Aoying'' whose main research keyword\footnote{
Authors' research keywords are obtained from ``Google Scholar'' except for ``Zhou Aoying'', whose information is obtained from ``ACM Digital Library'' due to the lack of information in ``Google Scholar.''
} is ``Query Processing, whereas that discovered by~\proposed~is ``Ee-Peng Lim'' whose main research keyword is ``Data \& Text Mining'', which is more relevant to the main research topic of ``Jiawei Han.'' 
In the case of ``Christos Faloutsos'', BGRL and~\proposed~discovered ``Michael J. Pazzani'' and ``David Jensen'' as the top-1 similar authors, respectively, both of whose main research topic is ``Machine Learning.''
However, it was interesting to see that ``David Jensen'', who was discovered by~\proposed, actually co-authored two papers ~\cite{senator2013detecting,kumar2008social} with ``Christos Faloutsos'', even though this information was missing in the Co.CS dataset.
On the other hand, there was no co-authorship between ``Christos Faloutsos'' and ``Michael J. Pazzani'', who is the top-1 similar author discovered by BGRL.
Thus, we argue that \proposed~discovers meaningful knowledge that is not revealed explicitly in the given graph.

\begin{table}[t]
\caption{Case studies on Coauthor CS dataset.}
\vspace{-2ex}
\begin{subtable}{1.0\linewidth}
\centering
\footnotesize
    \caption{Case 1: Which author is the most similar?}
    \vspace{-1ex}
    \renewcommand{\arraystretch}{0.99}
    \begin{tabular}{c|c|c|c|c}
    \hline
                       \begin{tabular}[x]{@{}c@{}}Query \\Author\end{tabular}& Model & \begin{tabular}[x]{@{}c@{}}Top-1 \\Similar Author\end{tabular}         & \begin{tabular}[x]{@{}c@{}}\# Co-authored \\Papers\end{tabular} & Student? \\ \hline\hline
    {\multirow{2}{*}{\begin{tabular}[x]{@{}c@{}}Jiawei\\Han\end{tabular}}}
                    & BGRL  & Ke Wang          & 14     & \xmark \\ 
                    & \proposed  & Xifeng Yan       & 87     & \cmark \\ \hline
    {\multirow{2}{*}{\begin{tabular}[x]{@{}c@{}}Christos\\Faloutsos\end{tabular}}} 
                    & BGRL  & Tina Eliassi-Rad & 27     & \xmark \\
                    & \proposed  & Hanghang Tong    & 47     & \cmark \\ \hline
    \end{tabular}
    
    \label{tab:case1}
    \vspace{3ex}
\end{subtable}
\begin{subtable}{1.0\linewidth}
\centering
\footnotesize
    \caption{Case 2: Which author will be connected in the future?}
    \vspace{-1ex}
    \renewcommand{\arraystretch}{0.95}
    \resizebox{0.99\linewidth}{!}{
    \begin{tabular}{c|c|c|c|c}
    \hline
                       \begin{tabular}[x]{@{}c@{}}Query \\Author\end{tabular}& Model & \begin{tabular}[x]{@{}c@{}}Top-1 \\Similar Author\end{tabular}         & \begin{tabular}[x]{@{}c@{}}\# Co-authored \\Papers\end{tabular} & \begin{tabular}[x]{@{}c@{}}Research \\Keywords\end{tabular}  \\ \hline\hline
    {\multirow{2}{*}{\begin{tabular}[x]{@{}c@{}}Jiawei\\Han\end{tabular}}}        
                    & BGRL  & Zhou Aoying & 0     & \textcolor{red}{Query Processing} \\ 
                    & \proposed  & Ee-Peng Lim & 0     & \textcolor{blue}{Data \& Text Mining} \\ \hline
    {\multirow{2}{*}{\begin{tabular}[x]{@{}c@{}}Christos\\Faloutsos\end{tabular}}} 
                    & BGRL  & Michael J. Pazzani & 0     & Machine Learning \\ 
                    & \proposed  & David Jensen    & 2    & Machine Learning \\ \hline
    \end{tabular}}
    \vspace{-1ex}
    \label{tab:case2}
\end{subtable}
\end{table}

\section{Conclusion}
In this paper, we propose a self-supervised learning framework for graphs, named \proposed, which learns node representations such that the relationship among nodes is invariant to augmentations, i.e., augmentation-invariant relationship. By doing so,~\proposed~allows the node representations to vary as long as the relationship among the nodes is preserved.
\proposed~learns diverse global relational information among the nodes considering the overall graph structure, while learning fine-grained relationship among structurally close nodes.
We also present in-depth discussions on how~\proposed~achieves the best of both worlds of contrastive/non-contrastive methods by relaxing strict constraints of previous methods with relational information of graph-structured data. 
Extensive experiments demonstrate that \proposed~consistently outperforms existing state-of-the-art methods. Moreover,~\proposed~1) demonstrates robustness to less informative or noisy features, and 2) improves performance on low-degree nodes, verifying its practicality in real-world applications.

\smallskip
\noindent \textbf{Acknowledgement.} This work was supported by Institute of Information \& communications Technology Planning \& Evaluation (IITP) grant funded by the Korea government(MSIT) (No.2022-0-00157, 100\%).
\bibliographystyle{ACM-Reference-Format}
\balance
\bibliography{sample-base}

\end{document}